\documentclass[sigconf]{acmart}

\usepackage{booktabs}

\usepackage{hhline}
\usepackage{amsmath,amssymb,amsthm}
\usepackage{todonotes}

\usepackage{subcaption}
\usepackage{graphicx} 
\usepackage{bmpsize}
\usepackage[noend]{algpseudocode}
\usepackage{algorithm}
\usepackage{multirow}
\usepackage{xspace}
\usepackage{array}
\usepackage{url}
\usepackage{lipsum}
\usepackage[utf8]{inputenc}		
\usepackage{amssymb}
\usepackage{pifont}
\usepackage{listings}
\usepackage{pgfplots}
\usepackage{xcolor,colortbl}
\usepackage{diagbox}
\usepackage{venndiagram}

\usetikzlibrary{patterns}

\newcommand{\rowcol}{\rowcolor{black!5}}

\usepackage{tikz}
\definecolor{dark-green}{rgb}{0,0.5,0}
\usetikzlibrary{shapes.geometric,arrows,positioning}

\newcommand{\model}{M}

\newcommand{\ed}{{\mathit{Euc}}}
\newcommand{\dist}{{\mathit{dist}}}
\newcommand{\fur}{{\mathit{fur}}}
\newcommand{\ts}{{\mathit{ts}}}

\algrenewcommand\algorithmiccomment[2][\itshape]{{#1\hfill\(\triangleright\)
    #2}}
\algrenewcommand{\algorithmicrequire}{\textbf{Input:}}
\algrenewcommand{\algorithmicensure}{\textbf{Output:}}

\newtheorem{definition}{Definition}

\usepackage[draft,footnote,nomargin]{fixme}
\fxusetheme{color}

\pgfplotsset{compat=1.14}

\begin{document}

\title{Testing Monotonicity of Machine Learning Models}

\author{Arnab Sharma}
\affiliation{%
	\institution{Department of Computer Science\\
		Paderborn University}
	\city{Paderborn}
	\country{Germany}}
\email{arnab.sharma@uni-paderborn.de}	

\author{Heike Wehrheim}
\affiliation{%
	\institution{Department of Computer Science\\
		Paderborn University}
	\city{Paderborn}
	\country{Germany}}
\email{wehrheim@uni-paderborn.de}



\begin{abstract}
 
Today, machine learning (ML) models are increasingly applied in decision making. 
This  induces an urgent need for 
{\em quality assurance} of ML models with respect to (often domain-dependent) requirements. 
\emph{Monotonicity} is one such requirement. It specifies a software as ``learned'' by an ML algorithm  to give an increasing prediction with the increase of some attribute values. While there exist multiple ML algorithms for {\em ensuring} monotonicity of the generated model, 
approaches for {\em checking} monotonicity, in particular of {\em black-box} models, are largely lacking. 

In this work, 
we propose {\em verification-based testing} of monotonicity, 
i.e., the formal computation of test inputs on a white-box model via verification technology, 
and the automatic inference of 
this approximating white-box model from the black-box model under test. 
On the white-box model, the space of test inputs can be systematically explored 
by a directed computation of  test cases. 
 The empirical evaluation on 90 black-box models shows verification-based testing can outperform 
adaptive random testing as well as property-based techniques with respect to effectiveness and efficiency. 


\end{abstract}


\maketitle

\section{Introduction}\label{sec:introduction}

Today, machine learning (ML) is increasingly employed to take decisions previously made by humans. 
This includes areas as diverse as insurance, banking, law, medicine or autonomous driving. 
Hence, quality assurance of ML applications becomes of prime importance. 
Consequently, researchers have started to develop methods checking various sorts of requirements. 
Depending on the domain of application, such methods target safety, security, fairness, robustness or balancedness of 
ML algorithms and models (e.g., \cite{HuangKWW17, Carlini017, galhotra2017fairness, SharmaW19}). 

A requirement frequently expected in application domains is {\em monotonicity} with respect to dedicated attributes. 
Monotonicity requires an increase in the value of some attribute(s) to lead to an increase in the value of the prediction (class attribute). 
For instance, a loan-granting ML-based software might be required to give larger loans whenever the value of attribute ``income'' 
gets higher, potentially even when other attribute values are changed.  

Monotonicity requirements occur in numerous domains like economic theory (house pricing, credit scoring, insurance premium determination),
 medicine (medical diagnosis, patient medication) or jurisdiction (criminal sentencing).  
In particular, monotonicity is often a requirement for ML software making acceptance/rejection decisions as it supports justification of a 
decision (``she gets a larger loan because she has got a higher income'').  
However, even if the training data used to generate the predictive model is monotone, the ML software itself might not be~\cite{Potharst:2002:CTP:568574.568577}. 
Hence, there are today a number of specialized ML algorithms which provide learning techniques guaranteeing monotonicity constraints
on models (e.g.~\cite{TehraniCDH11,Potharst:2002:CTP:568574.568577,YouDCPG17}).

Less studied is, however, the {\em validation} of monotonicity constraints, i.e.,~methods for answering the following question: 
\begin{quote}
   Given some black-box predictive model, does it satisfy a given monotonicity constraint?
\end{quote}
The term ``black-box model'' states the independence on  the ML-technology employed in the model 
(i.e., the technique should be equally applicable to e.g.~neural networks, random forests or support vector machines). 
We aim at a {\em model-agnostic} solution. 

\smallskip
\noindent 
In this paper, we present the first approach for automatic monotonicity testing of black-box machine learning models. 
Our approach systematically explores the space of test inputs by 
(1) the inference of a white-box model {\em approximating} the black-box model under test (MUT), and 
(2) the computation of counter examples to monotonicity on the white-box model via established verification technology. 
We call this approach {\em verification-based testing}. 
The computed counter examples serve as starting points for the generation of further test inputs by variation. 
If confirmed in the black-box model, they get stored as counter examples to monotonicity. 
If unconfirmed, they serve as input to an improvement of the approximation quality of the white-box model. 

More detailedly, our approach comprises the following key steps: 
\begin{itemize}
   \item {\bf White-box model inference.} A white-box model is generated by training a decision tree with data instances of the black-box model under test. 
   \item {\bf Monotonicity computation.} The decision tree is translated to a logical formula on which we use an SMT-solver for monotonicity 
      verification. 
   \item {\bf Variation.} The  computed counter examples are systematically varied (similar to strategies used in symbolic execution~\cite{King76}) 
     in order to increase the size of the test suite and 
     its coverage of the test space. 
   \item {\bf White-box model improvement.}  In case none of the counter examples are valid for the black-box model under test, 
     we employ the data instances together with the MUT's prediction to re-train the decision tree and thereby restart with an improved white-box model. 
\end{itemize}

We have implemented our approach and have experimentally evaluated it using standard benchmark data sets and 
both monotonicity aware and ordinary ML algorithms.  
Our experimental results suggest that our directed generation of test cases outperforms (non-directed) techniques like 
property-based testing wrt.~the effectiveness of finding monotonicity failures. 

Summarizing, this paper makes the following contributions: 
\begin{itemize}
  \item We formally define monotonicity of ML models.
  \item We present a novel approach to monotonicity testing via the usage of verification technology on an approximating white-box model. 
  \item We systematically evaluate our approach on 90 black-box models and compare it to state-of-the-art property-based 
    and adaptive random testing. 
  \item For the implementation of adaptive random testing as a baseline (to compare against), we design a distance metric specific 
    to monotonicity testing on ML models. 
\end{itemize} 

The paper is structured as follows. In the next section, we define monotonicity of machine learning models.  
In Section \ref{sec:approach} we describe verification-based testing and the way we have implemented adaptive random testing. 
Section \ref{sec:experiments} presents the results of our experimental evaluation. 
We discuss related work in Section \ref{sec:related} and conclude in Section \ref{sec:conclusion}.

\section{Monotonicity}\label{sec:monotone}

\begin{table}[t] \centering
   \caption{Example banking data set}
   \label{tab:bank-dataset}
   \begin{tabular}{l|ccc|l}
      No. & income & children & contract & loan \\
      \hline 
      1 & 100.0 & 1 & 20 & high \\
      2 & 25.0 & 0 & 2 & no \\
      3 & 17.8 & 3 & 5 & no \\
      4 & 25.5 & 2 & 15 & medium \\
      5 & 39.0 & 0 & 11 & medium \\
      \hline
   \end{tabular} 
\end{table} 

We start by introducing the basic terminology in machine learning and defining two notions of monotonicity.

A typical {\em supervised} machine learning (ML) algorithm works in two steps. 
Initially, it is presented with a set of data instances called {\em training data}. 
In the first (learning) phase, the ML algorithm generates a function (the {\em predictive model}), generalising from the training data by using some statistical techniques. The generated {\em predictive model} (short, model) is then used in the second (prediction) phase to predict classes for unknown data instances. 

Formally, the generated model is a function 
		 \[ \model: X_1 \times \ldots \times X_n \rightarrow Y \ ,\]
		 
\noindent where $X_i$ is the value set of {\em feature} $i$ (or attribute or characteristics $i$), $1 \leq i \leq n$, and $Y$ is the set of {\em classes}. We define $\vec{X}$ = $X_1 \times \ldots \times X_n$. 
The training data consists of elements from $\vec{X} \times Y$, i.e., data instances with known associated classes. 
During the prediction, the generated predictive model assigns a class $y\in Y$ to a data instance $(x_1, \ldots, x_n)\in \vec{X}$ 
(which is potentially not in the training data). We assume all $X_i$ and the set of classes $Y$ to be equipped with a total order $\preceq_i$ and $\preceq_Y$, respectively.

In this work, we check whether a given model is {\em monotone} with respect to a specific feature $i$. 

\begin{definition}\label{def:strong-mon} 
A model $M$ is  {\em strongly monotone}\footnote{Note that ``strong'' here does not refer to a strong increase in values, i.e., a definition with $\mathord{\prec}$ instead of $\mathord{\preceq}$.} with respect to a feature $i$ if for any two data instances $x = (x_1, \ldots, x_n)$, $x' = (x'_1, \ldots, x'_n)$ $\in \vec{X}$ we have 
$x_i \preceq_i x'_i$ implies $M(x) \preceq_Y M(x')$.
\end{definition}	 

Note that the feature values apart from the one with respect to which we are checking monotonicity can differ in an arbitrary way. 
Definition \ref{def:strong-mon} can be weakened as to only require an increasing prediction when all features values apart from the chosen one 
are kept.

\begin{definition}\label{def:weak-mon} A model $M$ is  {\em weakly monotone} with respect to a feature $i$ if for any two data instances $x=(x_1, \ldots, x_n), x'=(x'_1, \ldots, x'_n) \in \vec{X}$,  we have 
$(x_i \preceq_i x'_i) \land (\forall {j,j \neq i}.  x_j = x'_j)$ implies $ M(x) \preceq_Y M(x')$.
\end{definition}

\begin{figure}
\centering
\scalebox{.9}{\begin{tikzpicture}[level/.style={sibling distance=40mm/#1}, 
  inner/.style = {shape=rectangle, rounded corners,
    draw, align=center,  
    top color=blue!30, bottom color=blue!30},
  leaf/.style = {shape=rectangle,draw}]
  \node[inner] {contract}
    child { node[inner] {income} 
       child { node[leaf] {no} edge from parent node[above left,pos=.6] {$< 30$}}
       child { node[leaf] {high} edge from parent node[above right,pos=.6] {$\geq 30$} }
       edge from parent node[above left] {$< 10$} }
    child { node[inner] {income}
       child { node[leaf] {medium} edge from parent node[above left,pos=.6] {$< 50$}}
       child { node[leaf] {high} edge from parent node[above right,pos=.6] {$\geq 50$} }
       edge from parent node[above right] { $\geq 10$} };
\end{tikzpicture}}
\caption{A decision tree for the banking data set}
\label{fig:tree}
\vspace*{-.2cm}
\end{figure}

In the literature, the term monotonicity most often refers to our weak version. 
We also say that a training data set is strongly/weakly monotone if the above requirements hold for all elements in the set. 
As an example take the training data in Table~\ref{tab:bank-dataset} for a software making decisions about the granting of loans 
(inspired by~\cite{Potharst:2002:CTP:568574.568577}): The features of a person are the income (in thousand dollars), 
the (number of) children and the (duration of) current contract. 
The class ``loan'' can take four values: `no', `low', `medium' and `high', with total order `no' $<_Y$ `low' $<_Y$ `medium' $<_Y$ `high'. 
This data set is monotone\footnote{Note that we cannot apply our formal definitions here since the data set does not give us classes for all 
data instances.} for features ``contract'' and ``income'',  but not for feature ``children''. 

Figure \ref{fig:tree} gives a potential model (in the form of a decision tree) which the training on this data set could yield. 
It can correctly predict all instances in the training data. 
However, this model is not weakly monotone in feature ``contract'' anymore: Take e.g.~the following two data instances
\begin{center}
\begin{tabular}{lll}
   income=30.0, & children=0, & contract=9 \\
   income=30.0, & children=0, & contract=10 
\end{tabular}
\end{center}

\noindent While the prediction for the first instance is `high', it is `medium' for the latter.

We next define {\em group monotonicity} which extends Definitions~\ref{def:strong-mon} and \ref{def:weak-mon} to a set of features (called {\em monotone features}).

\begin{definition}\label{def:strong-group-mon} 
	A predictive model $M$ is said to be {\em strongly group monotone} with respect to a set of features $F=\{i_1,i_2,\ldots,i_m\} 
      \subseteq \{1, \ldots, n\}$ if for any two data instances $x=(x_1, \ldots, x_n), x'=(x'_1, \ldots, x'_n) \in \vec{X}$ we have  
$\forall j \in F: x_j \preceq_j x_j' $ implies $M(x) \preceq_Y M(x')$. 

Similarly, it is {\em weakly group monotone} with respect to $F$ if for all $x,x'$ we have 
$(\forall j \in F: x_j \preceq_j x_j' \wedge \forall j\notin F: x_j = x_j')$ implies $M(x) \preceq_Y M(x')$. 

\end{definition}

%


Strong group monotonicity sits in between weak and strong (single feature) monotonicity in that it allows some feature's values
to change in an arbitrary way while other values may only increase or stay as they are\footnote{Note that all $\mathord{\preceq}$
orders are reflexive.}. We see this as being a practically relevant case and have thus included it in our definitions. 

Finally, note that test cases for (both strong and weak) monotonicity are by these definitions 
{\em pairs} of data instances $(x,x')$, and during test execution these need to be checked 
for the property $M(x) \preceq_Y M(x')$. If the precondition of the respective monotonicity version holds for $x$ and $x'$ but 
$M(x) \not \preceq_Y M(x')$, we call the pair $(x,x')$ a 
{\em counter example} to monotonicity. 


\section{Testing approach}\label{sec:approach}
We conduct black-box testing to check monotonicity of the predictive model under test.   
Hence, in the following we assume the type of the MUT (i.e., which ML algorithm has been used for training)  to be unknown. 

\subsection{Adaptive random testing}

For the purpose of comparison, we have designed and implemented an {\em adaptive 
random testing} (ART)~\cite{DBLP:conf/asian/ChenLM04} approach for monotonicity testing 
which we describe first. 
Adaptive random testing aims at (randomly) computing test cases which are more evenly distributed 
among the test input space. To this end, it compares new candidates with the already 
computed test cases, and adds the one ``furthest'' away. 
The implementation of ``furthest'' requires the definition of a {\em distance metric}. 
For numerical inputs, this is often the Euclidean distance. 
For monotonicity, our test cases are however {\em pairs} $(x,x')$. 

Assume we are given two such pairs $(x,x')$ and $(z,z')$ and want to define how ``different'' 
they are.  
Assume furthermore that all the elements only contain {\em numerical} values\footnote{This can easily be 
achieved by some preprocessing step converting categorical to numerical values.}.  
 Every element $x=(x_1,\ldots,x_n)$ can then be considered to be a point in an $n$-dimensional 
space. We let $\ed(x,x')$ be the Euclidean distance between $x$ and $x'$, 
and $m_{x,x'}$ be the point laying at the middle of $x$ and $x'$.  
The metric which we employ captures two aspects:
we see two pairs $(x,x')$ and $(z,z')$ as being very different if 
(a) their Euclidean distances are far apart (e.g., $x$ is very close to $x'$, but $z$ far away from $z'$) 
and
(b) the middle of $(x,x')$ is far away from the middle of $(z,z')$. 
Formally, we define 

    \[ \dist\big((x,x'),(z,z')\big) = \frac{\lvert \ed(x,x') - \ed(z,z') \rvert}{2} + 
                                   \frac{\ed(m_{x,x'},m_{z,z'})}{2} 
                                     \]

\noindent Note that $\dist$ is positive-definite, symmetric and subadditive, i.e., indeed a metric. 
We use this distance function in the test case generation for ART within Algorithm \ref{alg:art}
(inspired by a definition of ART algorithms in \cite{DBLP:conf/dagstuhl/Walkinshaw16}). 

\begin{algorithm}[t] 
	\caption{$\mathit{artGen}$ (Test Generation for ART)}\label{alg:art}
	\begin{algorithmic}[1]
		\Require $F$ \Comment{set of monotone features}
		\Ensure  set of test cases
		\State $\ts$ := $\emptyset$; $count$ := 0;  
		\While{$count$ $<$ INI\_SAMPLES} \Comment{randomly generate start set}
		\State $x$ := random($\vec{X}$); 
		\State $x'$ := random($\{x' \mid \forall i\in F: x_i\preceq x'_i , 
                                                                        \forall j \notin F: x_j = x_j'\}$);  
		\If{$(x, x')\notin \ts$}
		\State $\ts$ := $\ts \cup \{(x, x')\}$; $count$++; 
             \EndIf
             \EndWhile
             \While{$\lvert \ts \rvert <$ MAX\_SAMPLES} \Comment{extend start set} 
                   \State $cand:=\emptyset$; $count$ :=0; 
                   \While{$count$ $<$ POOL\_SIZE}  \Comment{generate candidates}
		         \State $x$ := random($\vec{X}$); 
		         \State $x'$ := random($\{x' \mid \forall i\in F: x_i\preceq x'_i,
                                                                                 \forall j \notin F: x_j = x_j'\}$);  
		         \If{$(x, x')\notin cand$}
		         \State $cand$ := $cand \cup \{(x, x')\}$; $count$++; 
                      \EndIf
                  \EndWhile
                  \State $c_{\fur}$ := oneOf($cand$); \Comment{initialize with arbitrary cand.}
                  \State $maxDist$ := 0; 
                  \For{$c \in cand$} \Comment{determine ``furthest away'' cand.}
                      \State $dist$ := minDistance($c,\ts$); 
                      \If{$dist > maxDist$}
                           \State $c_{\fur}$ := $c$; $maxDist$ := $dist$; 
                      \EndIf
                  \EndFor 
                   \State $\ts$ := $\ts \cup \{c_{\fur} \}$; 
             \EndWhile 
		\State \Return $\ts$;  
	\end{algorithmic}
\end{algorithm} 

The first loop in Algorithm~\ref{alg:art} randomly computes a set of pairs $(x,x')$ to start with. 
Note that all these pairs already satisfy the precondition of --- in this case -- weak monotonicity 
by construction (line 4).
The second (outermost) loop extends this test set until it contains \texttt{MAX\_SAMPLES} pairs. 
It starts in line 9 by generating a set of candidates. The loop starting in line 16 then determines 
the candidate ``furthest away'' from the current test set $\ts$. It uses the function 
\texttt{minDistance} which computes the minimal distance between $c$ and elements of $\ts$ 
using the metric $\dist$. The furthest away candidate is put into the test set $\ts$ in line 20 and 
the algorithm returns with the entire test set in line 21. 
This test set is then subject to checking all pairs $(x,x')$ for monotonicity (not given as algorithm here).  

For checking strong monotonicity we follow a similar approach with the only exception being the 
generation of test input pairs $(x,x')$. In that case, the changes are in lines 4 and 11
 which become  
$x' := \text{random}(\{x' \mid $ $x \neq x', (\forall i \in F: x_i \preceq x_i')  \}$.

\subsection{Verification-based testing}

\begin{figure*}[t]
	\begin{center} 
		\includegraphics[width=.85\textwidth]{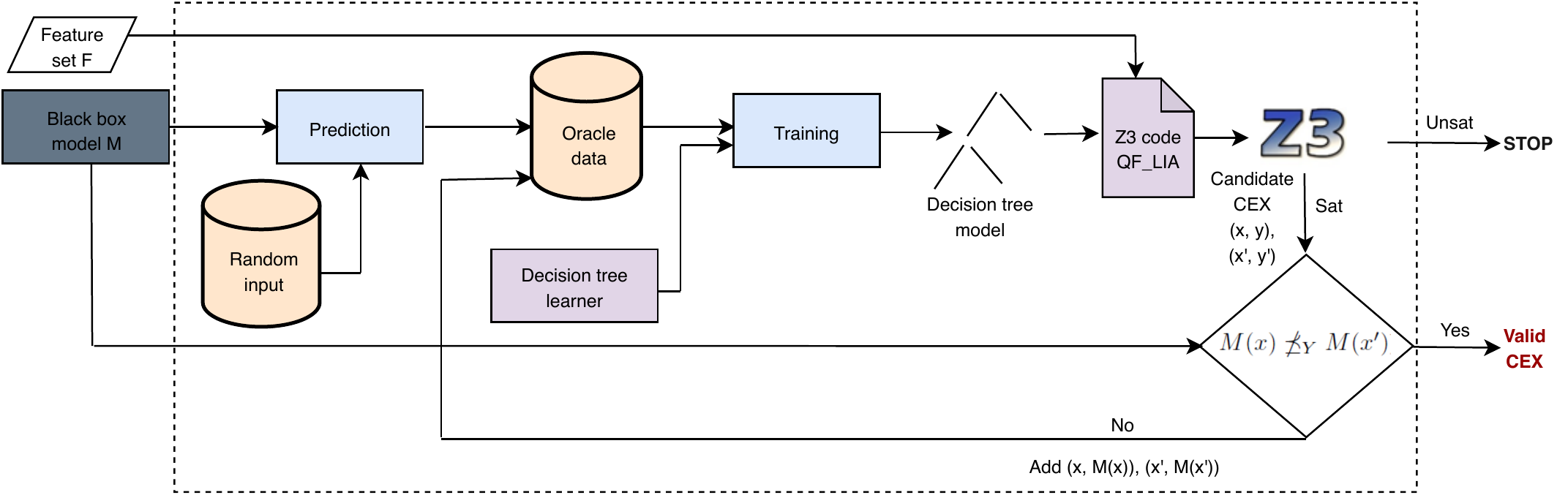}
	\end{center}
	\caption{Basic workflow of verification-based testing}
	\label{fig:workflow}
\end{figure*}

Next, we present our novel approach for the generation of test sets for 
monotonicity testing of a black-box model. 
The key idea therein is to approximate the MUT by a white-box model on which verification techniques can {\em compute} counter examples to 
monotonicity (in case these exist). These counter examples then serve as test inputs for the MUT, 
thereby achieving a target-oriented generation of test suites. 
We call this technique {\em verification-based testing} (VBT). 

The idea of approximating an unknown predictive model by a white-box model is already employed in the areas of 
interpretability of AI as well as testing of (non-ML) software. 
 In AI (Guidotti et al.~\cite{GuidottiMRTGP19}) an unknown black-box predictor is converted to an explainable model out of which explanations for humans can be constructed. 
In the field of testing, Papadopoulos and Walkinshaw~\cite{DBLP:conf/icse/PapadopoulosW15} use 
the inference of a predictive model from test sets to further extend this set.  None use it to compute 
counter examples to properties. 

In our work, we approximate the black-box MUT by a decision tree. The choice of it is driven by the desire to employ verification-based testing: counter examples to monotonicity are {\em automatically computable} for decision trees via SMT solvers. 
 Figure~\ref{fig:workflow} depicts the basic workflow; the exact interplay of components is detailed 
in Algorithm~\ref{alg:veriTest}.  Our approach is composed of four parts.

\textbf{White-box model inference.} The inputs to our approach are the predictive model $M$ (MUT) and a set of features $F$. 
In the first step, we train the decision tree. 
To this end, we generate so called {\em oracle data}, containing the predictions of the MUT for some randomly chosen input instances. This set of data instances are currently not used for testing monotonicity (but could be), rather they are employed for the purpose of generating the white-box model. A decision tree learner is then trained on the oracle data giving a decision tree model. 

 
\lstdefinelanguage{smt}{
     morekeywords={assert},
     morecomment=[l]{;}
}
\begin{figure}[t]
\begin{lstlisting}[frame=single, language=smt, 
tabsize=2,
columns=flexible,  
mathescape=true, commentstyle=\color{green!40!black}, keywordstyle=\color{blue}, 
linewidth =.97\linewidth,
%xleftmargin=.2\textwidth,
basicstyle=\footnotesize]
; Declaring components of x and x' and their classes
(declare-fun contract1 () Int)  (declare-fun income1 () Real)
(declare-fun contract2 () Int)  (declare-fun income2 () Real)
(declare-fun class1 () Int) (declare-fun class2 () Int)
; Specifying prediction of decision tree (no=0, medium=1, high=2)
(assert(=> (and (< contract1 10) (< income1 30)) (= class1 0)))
(assert(=> (and (< contract1 10) (>= income1 30)) (= class1 2)))
(assert(=> (and (>= contract1 10) (< income1 50)) (= class1 1)))
(assert(=> (and (>= contract1 10) (>= income1 50)) (= class1 2)))
(assert(=> (and (< contract2 10) (< income2 30)) (= class2 0)))
(assert(=> (and (< contract2 10) (>= income2 30)) (= class2 2)))
(assert(=> (and (>= contract2 10) (< income2 50)) (= class2 1)))
(assert(=> (and (>= contract2 10) (>= income2 50)) (= class2 2))) 
; Non-monotonicity constraint   
(assert (and (<= contract1 contract2) (= income1 income2)))
(assert (not (<= class1 class2)))
; Satisfiable? 
(check-sat)
;  Logical model extraction
(get-model)
\end{lstlisting}
\caption{Z3 code of the decision tree with strong monotonicity constraint}
\label{fig:z3code}
\vspace*{-.3cm}
\end{figure}

\textbf{Monotonicity computation.} Once we have generated the decision tree, the next step is to compute (non-)monotonicity. 
To this end, we use state of the art verification technology, namely the SMT solver Z3~\cite{MouraB08}. 
First, we translate the decision tree into a logical formula describing how the classes are predicted for inputs $x$ and $x'$. 
 Figure~\ref{fig:z3code} shows the Z3 code for the decision tree in Figure~\ref{fig:tree}. 
Therein, variables \verb+contract1+ and \verb+income1+\footnote{There is no variable for ``children''
since the decision tree is not using this feature.} describe test input $x$ and \verb+contract2+ and \verb+income2+ describe $x'$. 
The code also contains the non-monotonicity query for weak monotonicity wrt.~''income''. 
The last four lines of the code ask Z3 to check for satisfiability of all assertions and -- if yes (Sat) -- to return a logical model.
The logical model gives an evaluation for the variables such that all assertions are fulfilled. 
For our example, it can be found in Figure~\ref{fig:z3model}. 
It matches the counter example of Section~\ref{sec:monotone}. 

The counter example consists of a pair of data instances and their respective classes ($(x, y), (x', y')$) 
{\em as predicted by the decision tree}.  
 As the approximation of the MUT by the decision tree will typically be imprecise, this counter example might not be valid for the MUT: 
it is a {\em candidate} counter example (candidate CEX).  
Hence, we next check the validity of $M(x) \npreceq_Y M(x')$. If it holds, a true counter example to monotonicity of the MUT (valid CEX) has been found. 
If not, $(x,M(x))$ and $(x',M(x'))$ are added to the oracle data in order to increase precision of the approximation in later steps.   

When the output of Z3 is 'Unsat', we conclude that the generated decision tree is monotone wrt.~$F$ (but not necessarily the MUT $M$) and thus verification-based 
testing has been unable to compute a test case for non-monotonicity.

\textbf{Variation.} The basic workflow of Figure~\ref{fig:workflow} is complemented by {\em variation} techniques. 
Whenever we obtain a counter example which is not confirmed in the MUT, we need to make the approximating 
decision tree more precise. This is achieved by re-training. As this is a rather costly operation, we would like to avoid re-training for a single 
unconfirmed counter example and only re-train once we have collected a number of test pairs. 
This calls for a systematic generation of counter examples for which we need   Z3 to produce several different logical models 
for the same logical query. To this end, we employ two pruning techniques cutting off certain parts of the search space of Z3 
when computing logical models. 

\subsubsection{Pruning data instances}
Our first strategy is to call Z3 several times and simply disallow it to return the same counter example again. 
For our running example, we could simply add \verb+(assert (not (= contract1 9)))+ to our query and re-run the SMT solver. 
We can similarly do so for the values of the other features. 
This way we can often generate a large number of {\em similar} counter examples. 
E.g., for our example Z3 then returns an instance with \verb+contract1+ being 8, then 7, and so on. 
Algorithm \ref{alg:prunInst} describes how we generate new candidate pairs this way. 
Therein, name$_i$1 (name$_i$2) stands for the name of feature $i$ in instance $x$ ($x'$, respectively), 
e.g.\ \verb+income1+. 

The disadvantage of this approach is that Z3 will give counter examples considering only a few number of branches in the decision tree. Hence, a major part of the decision tree paths will remain unexplored. Next we define a second strategy which achieves better coverage of the tree and thereby an improved {\em test adequacy}. 

\begin{figure}[t]
\begin{lstlisting}[frame=single, tabsize=2, language=smt, 
columns=flexible, keywordstyle=\color{blue}, 
mathescape=true,
linewidth =.97\linewidth,
%xleftmargin=.2\textwidth,
basicstyle=\footnotesize]
     sat (model 
       (define-fun contract1 () Int 9)
       (define-fun income1 () Real 30.0)
       (define-fun class1 () Int 2)
       (define-fun contract2 () Int 10)
       (define-fun income2 () Real 30.0)
       (define-fun class2 () Int 1))
\end{lstlisting}
\caption{Logical model for the query of Fig.~\ref{fig:z3code}}
\label{fig:z3model}
\vspace*{-.3cm}
\end{figure}

\begin{algorithm}[t] 
	\caption{$\mathit{prunInst}$ (Pruning data instances)}\label{alg:prunInst}
	\begin{algorithmic}[1]
		\Require  $(x, x')$  \Comment{A candidate pair}
		\Statex \hspace*{.4cm} $\varphi$ \Comment{Logical formula}
		\Ensure set of candidate pairs
             \State cand-set:= $\emptyset$; 
		\For{$i := 1$ to $n$} \Comment{$n$: number of features} 
                 \State $\psi$ := $\varphi \wedge \neg ($name$_i1 = x_i)$;
                 \If {SAT($\psi$)}
                      \State cand-set := cand-set $\cup$ get-model($\psi$);
                 \EndIf
             \EndFor	
             \For{$i := 1$ to $n$}  
                 \State $\psi$ := $\varphi \wedge \neg ($name$_i2 = x'_i)$;
                 \If {SAT($\psi$)}
                      \State cand-set := cand-set $\cup$ get-model($\psi$);
                 \EndIf
             \EndFor			 
		\State \Return cand-set; 		
	\end{algorithmic}

\end{algorithm} 

\begin{algorithm}[t] 
	\caption{$\mathit{prunBranch}$ (Pruning branches)}\label{alg:prunBranch}
	\begin{algorithmic}[1]
		\Require  $(x, x')$  \Comment{A candidate pair}
             \Statex \hspace*{.4cm} tree \Comment{Decision tree} 
		\Statex \hspace*{.4cm} $\varphi$ \Comment{Logical formula}
		\Ensure set of candidate pairs
             \State cand-set:= $\emptyset$; 
             \State $(c_1, \ldots, c_m)$ := getPath(tree,$x$); \Comment{Path of $x$ in tree}
		\For{$i := 1$ to $m$} \Comment{toggle path conditions}  
                 \State $\psi$ := $\varphi \wedge \neg c_i$;
                 \If {SAT($\psi$)}
                      \State cand-set := cand-set $\cup$ get-model($\psi$);
                 \EndIf
             \EndFor	
             \State $(c_1, \ldots, c_k)$ := getPath(tree,$x'$); \Comment{Path of $x'$ in tree}
 		\For{$i := 1$ to $k$} \Comment{toggle path conditions}  
                 \State $\psi$ := $\varphi \wedge \neg c_i$;
                 \If {SAT($\psi$)}
                      \State cand-set := cand-set $\cup$ get-model($\psi$);
                 \EndIf
             \EndFor		 
		\State \Return cand-set; 		
	\end{algorithmic}
\end{algorithm} 


\subsubsection{Pruning branches}
In this case we use a global approach to traverse as many paths as possible. Once a   test pair $(x,x')$ is found, 
we identify the paths in the tree which this pair takes. Then we 
toggle the conditions on $x$'s path and on $x'$'s path, one after the other. 
Other toggling strategies are possible.
This ``condition toggling'' follows the established strategy of determining path conditions in 
symbolic execution and systematically negating conditions for concolic testing~\cite{GodefroidKS05, SenMA05}. 
Figure~\ref{fig:prune} illustrates it on one path (the conditions in bold
 are added to the Z3 code) and Algorithm~\ref{alg:prunBranch} gives the algorithm.  

\begin{figure}[t] 

\begin{subfigure}{.2\linewidth}
\begin{tikzpicture}[level/.style={sibling distance=10mm, level distance=10mm}, 
  inner/.style = {shape=rectangle, rounded corners,
    draw, align=center,  baseline, 
    top color=blue!30, bottom color=blue!30},
  leaf/.style = {shape=rectangle}]
  \node[inner] {\ }
    child { node[inner] {\ }
       child { node[inner] {\ } 
          child { node[leaf] {} edge from parent node[above left] {$c_3$} }
          child { node[leaf] {} edge from parent[dotted] {} } 
         edge from parent node[above left] {$c_2$}}
       child { node[leaf] {} edge from parent[dotted]  }
    edge from parent node[above left] { $c_1$} }
    child { node[leaf] {} edge from parent[dotted] };
\end{tikzpicture}
\end{subfigure} \hfill 
\begin{subfigure}{.2\linewidth}
\begin{tikzpicture}[level/.style={sibling distance=10mm, level distance=10mm}, 
  inner/.style = {shape=rectangle, rounded corners,
    draw, align=center,  baseline, 
    top color=blue!30, bottom color=blue!30},
  leaf/.style = {shape=rectangle}]
  \node[inner] {\ }
    child { node[inner] {\ }
       child { node[inner] {\ } 
          child { node[leaf] {} edge from parent[dotted] {} } 
          child { node[leaf] {} edge from parent node[above right] {$ \mathbf{\neg c_3}$} }
         edge from parent node[above left] {$c_2$}}
       child { node[leaf] {} edge from parent[dotted]  }
    edge from parent node[above left] { $c_1$} }
    child { node[leaf] {} edge from parent[dotted] };
\end{tikzpicture}
\end{subfigure}  \hfill 
\begin{subfigure}{.2\linewidth} 
\begin{tikzpicture}[level/.style={sibling distance=10mm, level distance=10mm}, 
  inner/.style = {shape=rectangle, rounded corners,
    draw, align=center,  baseline, 
    top color=blue!30, bottom color=blue!30},
  leaf/.style = {shape=rectangle}]
  \node[inner] {\ }
    child { node[inner] {\ }
       child { node[leaf] {} edge from parent[dotted]  }
       child { node[leaf ] {\ } 
         edge from parent node[above right] {$\mathbf{\neg c_2}$}}
    edge from parent node[above left] { $c_1$} }
    child { node[leaf] {} edge from parent[dotted] };
\end{tikzpicture}
\end{subfigure}\hfill 
\begin{subfigure}{.2\linewidth} 
\begin{tikzpicture}[level/.style={sibling distance=10mm, level distance=10mm}, 
  inner/.style = {shape=rectangle, rounded corners,
    draw, align=center,  baseline, 
    top color=blue!30, bottom color=blue!30},
  leaf/.style = {shape=rectangle}]
  \node[inner] {\ }
       child { node[leaf] {} edge from parent[dotted]  }
    child { node[leaf] {} edge from parentnode[above right] {$\mathbf{\neg c_1}$} };
\end{tikzpicture}
\end{subfigure}
\caption{Illustration of condition toggling}
\label{fig:prune}
\vspace*{-.5cm}
\end{figure} 


\smallskip
\noindent 
While the first pruning strategy tries to find new pairs in the local neighbourhood of a counter example, the second strategy globally 
searches for counter examples and thus achieves a better coverage. 
Algorithm~\ref{alg:veriGen} summarizes one such pass of counter example candidate generation from the decision tree. 

\textbf{White-box model improvement.} Once we have collected a larger set of test pairs (all candidate counter examples), 
we test whether they are also counter examples to monotonicity for the MUT $M$. 
We furthermore check whether $M$'s prediction differs from the decision tree's prediction on the candidates. 
If yes, they will be added to the oracle data to re-train the decision tree.

\textbf{Overall algorithm.}
Algorithm~\ref{alg:veriTest} summarizes all these steps in one algorithm. 
It interleaves generation of test cases with monotonicity checking, i.e., after one call to $\mathit{veriGen}$ it first checks 
all candidates and only if none of these are counter examples to monotonicity, it starts retraining. 
 Two constants play a r\^{o}le in this algorithm: 
the constant \texttt{MAX\_ORCL} fixes the number of data instances to be generated for 
initially training the decision tree.
and (again) constant \texttt{MAX\_SAMPLES} 
limits the number of samples we 
generate\footnote{Note that due to variation, $\ts$ might become slightly larger than \texttt{MAX\_SAMPLES}.}. 

\begin{algorithm}[t] 
	\caption{$\mathit{veriGen}$ (Test Generation for VBT)}\label{alg:veriGen}
	\begin{algorithmic}[1]
		\Require $F$ \Comment{Set of features}
             \Statex \hspace*{.4cm} tree \Comment{trained decision tree}          
		\Ensure set of test cases
             \State $cs$ := $\emptyset$;  
                 \State $\varphi$ := tree2Logic(tree); 
                 \State $\varphi$ := $\varphi \wedge {} $nonMonConstr($F$);
                 \If {UNSAT($\varphi$)} \Comment{No CEX cand.\ found}
                     \State \Return $\emptyset$;          
                \EndIf
                \State $((x,y),(x',y'))$ := getModel($\varphi$); \Comment{Gen.\ candidates}
                \State $cs:=\{((x,y),(x',y'))\}$; 
                \State $cs:=cs \cup {}   $prunInst$(x,x',\varphi)  
                                    \cup {}  $prunBranch$(x,x'$,tree,$\varphi)$;
		    \State \Return $cs$; 
	\end{algorithmic}
\end{algorithm} 

\begin{algorithm}[t] 
	\caption{$\mathit{veriTest}$ (Verification-based testing)}\label{alg:veriTest}
	\begin{algorithmic}[1]
		\Require  $M$ \Comment{Model under test}
		\Statex \hspace*{.4cm} $F$ \Comment{Set of features}         
		\Ensure counter example to monotonicity or empty
             \State $orcl\_data$ := $\emptyset$; $\ts$ := $\emptyset$; $cs$ := $\emptyset$; 
             \While{$\lvert orcl\_data \rvert <$ MAX\_ORCL}  
                  \State $x$:= random($\vec{X}$);
                  \If {$(x,M(x)) \notin orcl\_data$}
                      \State $orcl\_data := orcl\_data \cup \{(x,M(x))\}$;
                  \EndIf 
             \EndWhile 
             \While {$\lvert \ts \rvert <$ MAX\_SAMPLES}
                 \State tree := trainDecTree($orcl\_data$);
                 \State $cs$ := veriGen($F$,tree); 
                \For {$((x,y),(x',y')) \in cs$} \Comment{Duplicates?}
                     \If {$(x,x') \notin \ts$}
                         \State $\ts := \ts \cup  \{(x,x')\}$;
                     \Else
                         \State $cs := cs \setminus \{((x,y),(x',y'))\}$; 
                     \EndIf 
                \EndFor 
                \If {$cs = \emptyset$} \Comment{No new candidates?}
                    \State \Return Empty;
                \EndIf
                \For {$((x,y),(x',y')) \in cs$}
                   \If {$M(x) \npreceq_Y M(x')$} \Comment{Valid CEX?}
                       \State \Return $((x,y),(x',y'))$
                   \EndIf
                   \If {$y \neq M(x)$} \Comment{Different prediction?}
                       \State $orcl\_data:=orcl\_data \cup \{(x,M(x))\}$;  
                   \EndIf
                   \If {$y' \neq M(x')$}
                       \State $orcl\_data:=orcl\_data \cup \{(x',M(x'))\}$;  
                   \EndIf
                \EndFor
             \EndWhile 
		\State \Return Empty; \Comment{No counter example found}
	\end{algorithmic}
\end{algorithm} 
 
\section{Evaluation} \label{sec:experiments}


We have implemented adaptive random as well as verification-based testing for monotonicity in Python and 
have comprehensively evaluated VBT. 
The following research questions guided our evaluation. 
We broadly divide these questions into two categories. The first category concerns the comparison of VBT with existing 
techniques with respect to effectiveness and efficiency. 
The second category evaluates the performance of VBT itself. 


\begin{enumerate}
	\item[\textbf{RQ1.}]  {\bf Effectiveness} \\
How does VBT compare to existing testing approaches with respect to the error detection capabilities? 
	\item[\textbf{RQ2.}] {\bf Efficiency} \\
How does VBT compare to existing testing approaches with respect to the effort for error detection?
\end{enumerate}

    
To analyse VBT itself we have focused on the following research questions:

\begin{enumerate}
	\item[\textbf{RQ3.}]  {\bf Approximation quality} \\
Can the decision trees adequately represent black-box models?
      \item[\textbf{RQ4.}] {\bf Strategy selection} \\ 
Which pruning strategy performs better in computing non-monotonicity?
\end{enumerate} 

\smallskip
\noindent 
We have carried out the following experiments to evaluate these research questions.

\textbf{RQ1.} As there are no specific approaches for computing monotonicity of a given black-box model, we use property-based testing (i.e., a variant of QuickCheck~\cite{DBLP:conf/icfp/ClaessenH00} for Python) and (our own implementation of) adaptive random testing as baseline approaches to compare against. Our intention is to measure how well a technique is able to generate test cases revealing non-monotonicity of a given model. 
To this end, we have taken 8 ML algorithms from the state of the art ML library \verb+scikit-learn+ plus one monotonicity aware classifier~\cite{light}, and trained them on ten data sets 
(see below) to generate 90 different predictive models.  
For these models it is first of all unknown whether they are monotone or not, so we lack a ground truth. 
The comparison is thus performed on the basis of just counting the number of models in which non-monotonicity is detected\footnote{Note that none of the techniques produce false positives since they all perform a dynamic analysis.}. 
We perform the evaluation for weak (group) monotonicity as it is the standardly employed concept.

%

\begin{table}\centering
	
	\small
	\caption{Data sets and their characteristics}
	\label{tab:datasets}
	\begin{tabular}{r r r r r}
		\toprule
		\textbf{Name} & \textbf{\#Features} & \textbf{\#Group} & \textbf{\#Instances} & \textbf{\#TreeNodes}\\
		\midrule
		\rowcol \textit{Adult} & 13 & 4 & 32561 & 4673  \\
		\textit{Diabetes} & 8 & 5 & 768 & 267 \\
		\rowcol \textit{Mammographic} & 6 & 3 & 961 & 481 \\
		\textit{Car-evaluation} & 6 & 4 & 1728 & 167 \\
		\rowcol \textit{ESL} & 4 & 2 & 488 & 295 \\
		\textit{Housing} & 13 & 3 & 506 & 107 \\
		\rowcol \textit{Automobile} & 24 & 10 & 205 & 53  \\
		\textit{Auto-MPG} & 7 & 5 & 392 & 117 \\
		\rowcol \textit{ERA} & 4 & 2 & 1000 & 87 \\
		\textit{CPU} & 6 & 5 & 209 & 11 \\
		
		\bottomrule
	\end{tabular}
\end{table}

\textbf{RQ2.} To answer RQ2, we have carried out experiments in the same setting as considered for RQ1. We wanted to evaluate how efficient our verification-based testing approach is compared to adaptive random and property-based testing. 
To this end, we (a) determine the run time needed for test generation and checking, and 
(b) the number of generated test cases necessary for finding the first error or, in case of a failure in error detection, just the number 
of generated tests. In the latter case, this is the maximal number of samples to be considered by the approach, 
which is configurable for property-based testing and which is \texttt{MAX\_SAMPLES} for ART and VBT (see below for values used). 

 
\textbf{RQ3.} As we employ decision trees for computing candidate counter examples, the performance of VBT crucially depends on 
decision trees to adequately approximate the black-box model. ``Adequately'' here means adequate for the task of computing counter 
examples. In general, the decision tree model and the black-box will differ on some predictions. 
The inadequacy of the decision tree shows up whenever we need to re-train it several times in order to find a proper counter example. Hence, for RQ3 we 
determine the number of re-trainings for all 90 models and weak monotonicity.

\textbf{RQ4.} For test generation (Algorithm \textit{veriGen}) we have implemented two different pruning strategies to achieve better coverage of the decision tree. So we wanted to find out which strategy is better in terms of finding counter examples. 
For the evaluation, we slightly change the setting. First, instead of only using weak monotonicity, we also check for strong monotonicity 
since we had the impression in initial experiments that the pruning strategies might behave differently for the weak and strong version.
Second, we modify VBT such that it generates several counter examples (simply by not stopping it on the first one) and 
compute the achieved {\em detection rate}, i.e.~number of detected errors divided by number of overall test cases 
$\frac{\# \text{errors}}{\# \text{test cases}}$.

\subsection{Setup}
We have collected our 10 data sets from the UCI machine learning repository\footnote{https://archive.ics.uci.edu/ml} and the OpenML data repository\footnote{https://www.openml.org}.
These training data sets have also been used in existing works~\cite{KotlowskiS09, TehraniCDH11} on monotonicity. Table \ref{tab:datasets} shows the data sets and their characteristics, i.e., the number of features, the size of the group (number of features in the group) and the number of data instances in the set. 
We have also computed the number of nodes of the decision tree model (column \textit{\#TreeNodes}) 
when being trained on the corresponding dataset 
to give a rough idea about the size of decision trees generated in VBT.   

The {\em monotone features}
are chosen based on our own assumptions about the domain and previous works. For instance, in case of the Adult data set, where a model predicts whether a person's income is at least 
\$50,000, we check monotonicity with respect to the group consisting of age, weekly working hours, capital-gain and education level~\cite{YouDCPG17}. 

The eight classification algorithms which have been taken from \verb+scikit-learn+ are 
kNN, Neural Networks (NN), Random Forests (RF), Support Vector Machines (SVM), Naive Bayes (NB), AdaBoost, GradientBoost and Logistic Regression. We have used a linear kernel for SVM and a Gaussian version of NB for 
our experiments. There are several other ML algorithms which can be found in the library but the eight algorithms chosen here belong to the most basic family of ML classifiers.  
The 9th ML classifier is the monotonicity aware algorithm LightGBM~\cite{light} which has been specifically designed to construct models being monotone with respect to a given set of features. The monotonicity constraints can be enforced during the training phase of this algorithm. This should guarantee monotonicity but -- as initial experiments have revealed -- LightGBM does not entirely manage to rule out non-monotonicity. This classifier is an excellent benchmark for the three approaches since there are only a very few erroneous input pairs and the challenge is to generate exactly these as test cases.

%

We have evaluated the accuracy score while generating predictive models and used the score to adjust the hyperparameters of the learning algorithms. The input parameters of \emph{artGen} and \emph{veriTest} algorithms have been chosen based on execution time of some initial experiments. The parameter \texttt{POOL\_SIZE} of the 
$artGen$ algorithm is set to half of \texttt{INI\_SAMPLES} which is 100; for \texttt{MAX\_SAMPLES} we use 1000. 

We have created oracle data in the verification-based testing approach by generating random data instances (90\%) and also taking training data instances randomly (10\%)\footnote{Note that for simplicity the stated algorithm does not include the training part.}. This choice is influenced by the work of Johansson et al.~\cite{JohanssonN09} who found that using random data instances to approximate a model gives the best result. 

We have used \textsc{Hypothesis} \cite{hypothesis}, a Python version of QuickCheck \cite{DBLP:conf/icfp/ClaessenH00} as our property-based testing tool.
Property-based testing allows users to specify the property to be tested. 
The parameters of this tool have been set in accordance with the \emph{artGen} and \emph{veriTest} algorithms (parameter for upper 
bound of test cases is 1000 like  \texttt{MAX\_SAMPLES}). 

Finally, because all three approaches involve some sort of randomness, every experiment was carried out ten times. The results give the 
arithmetic mean over these ten runs. 
The experiments were run on a machine with 2 cores Intel(R) Core(TM) i5-7300U CPU with 2.60GHz and 16GB memory using Python version 3.6. 

\subsection{Results}

Next, we report on the findings of our experiments while evaluating the research questions. 

\textbf{RQ1 - Effectiveness.} Table~\ref{tab:failuresCount} shows the results of the experiments for RQ1. 
It gives the number of models (out of 90 in total) for which our approach verification-based testing (VBT) and the two baselines 
adaptive random testing (ART) and property-based testing (PT) were able to detect non-monotonicity. 
Note again that the ground truth, i.e., which models are in fact non-monotone, is unknown, but all reported non-monotonicity cases 
are true positives. Per classifier 10 models were tested. 

\begin{table}[t]\centering
	\small
	\caption{Number of non-monotonicity detections}
	\label{tab:failuresCount}
	\begin{tabular}{c c c c}
		\toprule
		\textbf{Classifiers} & \textit{VBT} & \textit{ART} & \textit{PT}\\
		\midrule
		\rowcol k-NN & 9 & 9 & 7  \\
		Logistic Regression & 8 & 8 & 6\\
		\rowcol Naive Bayes & 7 & 4 & 5 \\
		SVM & 9 & 8 & 5 \\
		\rowcol Neural Network & 8 & 6 & 4 \\
		Random Forest & 9 & 9 & 5 \\
		\rowcol AdaBoost & 8 & 7 & 5 \\
		GradientBoost & 8 & 7 & 5 \\
		\rowcol LightGbm & 2 & 0 & 0 \\
		\bottomrule
		Overall & 68 & 58 & 42 \\
		\bottomrule
	\end{tabular}
	\vspace{-0.1cm}
	
\end{table}

The results show that verification-based testing is more effective in detecting non-monotonicity than both adaptive random and property-based testing. It also shows that adaptive random testing can outperform (pure) property-based testing because 
-- with the help of the distance metric -- it more systematically 
generates test cases covering the test input space. 

Another interesting result is the detection of non-monotonicity in two (out of 10) models generated by LightGBM (and the fact that ART and PT fail to detect it). 
LightGBM is a monotonicity-aware classifier which is supposed to just generate monotone models. 
For two of the training sets it has however failed to do so, resulting in a model which still has a small number of non-monotone pairs 
which VBT can find, but ART and PT cannot. 

Given these differences in numbers, we also wanted to know whether the non-monotonicity detections of ART and PT are simply a subset 
of those of VBT. This is actually not the case. The Venn diagram in Figure~\ref{fig:venn} shows the distribution of counter examples 
onto the three approaches. 26 models are in the intersection of all three techniques. 
For the rest, only one or two of the approaches could detect non-monotonicity. 
The diagram also shows that there are 9 models for which both PT or ART can detect non-monotonicity,  
but VBT cannot. 
These models are all models trained on the ERA (6 models) and ESL (3 models) data sets. Looking at the models themselves, 
it turns out that their accuracy (wrt.~the training data) is always very low (below 0.62 and for ERA even below 0.3). 
Hence, it seems that generalization from this training data is difficult for ML algorithms, and the decision trees in VBT seem to generalize 
in a different way than the black-box models and hence approximate them less well. This also shows in the results of RQ3 below 
concerning the ERA and ESL data sets.  

\begin{figure}[t] 
 
\begin{tikzpicture}
    \tikzset{venn circle/.style={draw=gray,text opacity=2,fill opacity=0.2,circle,minimum width=3cm,fill=#1,line width=1.5pt}}
    \begin{scope}[blend mode=screen]
          \node [venn circle = yellow] (A) at (0,0) {};
          \node [label] (A)  at (-1.5,-1.5) {\textbf{ART}};
          \node [venn circle = green] (B) at (2,0) {};
          \node [label] (B) at (3.5,-1.5) {\textbf{PT}};  
          \node [venn circle = orange] (C) at (1,1.5) {};
          \node [label] (C) at (1,3.3) {\textbf{VBT}};

         \node at (1,0.55) {\textbf{26}};
         \node at (0.2, 0.85) {\textbf{24}};
         \node at (1.8, 0.85) {\textbf{8}};
         \node at (1, 1.85) {\textbf{10}};
         \node at (-0.4, -0.1) {\textbf{1}};
         \node at (1, -0.55) {\textbf{7}};
         \node at (2.2, -0.1) {\textbf{1}};
         
        \end{scope}
\end{tikzpicture}
 
   \caption{Venn diagram showing distribution of detected non-monotone models on approaches}
   \label{fig:venn}
   \vspace{-0.1cm}
\end{figure}

\smallskip
\noindent 
Summarizing the findings of RQ1 in our experiments, we get 

\begin{center}
\fcolorbox{blue}{blue!10}{\parbox{0.85\columnwidth}{On average, VBT is more effective 
than ART and PT in detecting non-monotonicity of black-box ML models.}}
\end{center}

\bigskip
 


\noindent \textbf{RQ2 - Efficiency.} For RQ2, we designed experiments to evaluate how efficient VBT is in comparison to ART and PT.  Figure~\ref{fig:execution-time} shows the runtime of 
the three approaches for testing monotonicity. The x-axis enumerates the 90 tasks (i.e., models to be tested) where 
the tasks are sorted in ascending order of runtime per approach, and the y-axis gives the runtime for testing the task (in seconds, on a logarithmic scale). 
It shows that for all tasks VBT takes less time than both ART and PT even though our approach consists of several steps (including the training of a decision tree) to compute monotonicity. 

Adaptive random testing takes (on average) the same amount of time for all tasks. During ART, most of the time is needed for creating the input test cases which are ``furthest'' away from each other. As the size of test inputs is always the same for all the test cases, the time does not vary that much. On the other hand, property-based testing performs better than ART in most of the cases apart from some exceptions. 

\pgfplotsset{
	compat=1.14,
	width=\columnwidth,
	height=6cm
}

\begin{figure}[t]
	\begin{center}
		\begin{tikzpicture}
		\definecolor{clr1}{rgb}{0.64,0.64,0.82}
		\definecolor{clr2}{rgb}{0.0,0.0,0.0}
		\definecolor{clr3}{rgb}{0.21,0.27,0.31}
		\begin{axis}[
		legend pos=north west, 
		ytick = {1,10,100,1000,10000},
		ylabel={runtime (in seconds)},
		ymode = log, 
		xtick = {10,20,30,40,50,60,70,80,90}, 
		xmin=10, xmax=90,
		ymin = 1, ymax=10000,
		xticklabels = {1, 20, 30, 40, 50,60,70,80, 90}, 
		xlabel = {Tasks}
		]
		\addplot
		[clr2, line width=0.3mm, dotted] coordinates{
			(1,94.1) (2,94.5) (3,95.9) (4,95.98) (5,97.1) (6,97.6) (7,99.4) (8,99.47) (9,103.6) (10,104.2) (11,107.5) (12,108.5) (13,109.5) (14,109.7) (15,109.9) (16,114.3) (17,115.4) (18,121.4) (19,128.2) (20,147.8) (21,150.6) (22,152.99) (23,153.12) (24,153.2) (25,158.78) (26,173.4) (27,173.8) (28,179.1) (29,216.3) (30,222.8) (31,223.6) (32,225.9) (33,226.2) (34,227.2) (35,227.5) (36,229.6) (37,229.9) (38,230.6) (39,231.1) (40,232.4) (41,232.5) (42,237.3) (43,240.1) (44,242.8) (45,248.7) (46,259.1) (47,260.8) (48,261.5) (49,261.9) (50,273.9) (51,278.9) (52,329.9) (53,335.6) (54,338.9) (55,340.8) (56,373.2) (57,385.1) (58,387.8) (59,410.5) (60,412.1) (61,412.4) (62,413.2) (63,414.3) (64,415.7) (65,416.2) (66,416.9) (67,419.6) (68,421.1) (69,421.6) (70,421.9) (71,423.4) (72,424.02) (73,434.44) (74,435.6) (75,436.8) (76,461.55) (77, 463.5) (78, 464.6) (79, 468.7) (80, 469.9) (81, 475) (82, 476.2) (83, 477.3) (84, 478.1) (85, 479.5) (86, 480.2) (87, 481.5) (88, 481.7) (89, 481.9) (90, 490)
		};
		
		\addplot
		[clr1, line width=0.3mm, dashed] coordinates{
			(1,8.4) (2,8.5) (3,8.56) (4,8.6) (5,8.95) (6,9.07) (7,9.1) (8,9.26) (9,9.56) (10,9.68) (11,9.96) (12,10.27) (13,10.9) (14,10.98) (15,11.19) (16,11.6) (17,11.8) (18,12.15) (19,13.95) (20,14.02) (21,14.37) (22,15.64) (23,16.6) (24,16.69) (25,17.14) (26,17.19) (27, 17.28) (28,17.75) (29,19.48) (30,19.75) (31,20.53) (32,24.64) (33,26.19) (34,26.31) (35,30.8) (36,32.26) (37,35.88) (38,39.21) (39,41.94) (40,45.1) (41,45.86) (42,46.78) (43,47.1) (44,50.22) (45,50.52) (46,55.91) (47,60.93) (48,62.31) (49,62.79) (50,65.03) (51,71.34) (52, 72.3) (53,73.5) (54,73.8) (55,74.1) (56,74.5) (57,74.9) (58,76.1) (59,77.2) (60,78.4) (61,78.8) (62,79.14) (63,80.38) (64,93.7) (65,96.74) (66,96.77) (67,112.14) (68,113.47) (69,115.61) (70,121.6) (71,123.75) (72,123.92) (73,126.88) (74,137.69) (75,138.4) (76,139.58) (77,144.59) (78,156.54) (79,173.69) (80,184.47) (81,263.42) (82,413.64) (83,451.52) (84,541.32) (85,580.17) (86,673.3) (87,988.44) (88, 3980.2) (89,5433.12) (90,8567)
		};

		\addplot
		[clr3, line width=0.3mm] coordinates {
			(1,1.08) (2,1.45) (3,1.47) (4,1.54) (5,1.56) (6,1.68) (7,1.77) (8,2.26) (9,2.5) (10,2.53) (11,3.09) (12,3.17) (13,3.21) (14,3.34) (15,3.39) (16,3.41) (17,3.54) (18,3.61) (19,3.63) (20,3.66) (21,3.69) (22,3.72) (23,4) (24,4.1) (25,4.17) (26,4.2) (27,4.24) (28,4.25) (29,4.26) (30,4.34) (31,4.51) (32,4.52) (33,4.6) (34,4.75) (35,4.78) (36,4.83) (37,4.84) (38,4.85) (39,4.86) (40,5.58) (41,5.61) (42,5.86) (43,5.99) (44,6.16) (45,6.42) (46,6.43) (47,6.89) (48,6.99) (49,7.1) (50,7.38) (51,7.9) (52,8.3) (53,8.59) (54,8.67) (55,8.74) (56,8.96) (57, 8.99) (58, 9.01) (59, 9.12) (60, 9.39) (61, 9.48) (62, 10.42) (63, 10.67) (64, 10.8) (65, 10.87) (66, 12.4) (67,14.8) (68,14.9) (69,15.5) (70,18.1) (71,18.5) (72,26.6) (73, 28.6) (74,31.3) (75,39.1) (76,42.2) (77,49.3) (78,50.9) (79,52.07) (80,56.5) (81,86.4) (82,94.2) (83,100.3) (84,102.2) (85,131) (86,150.6) (87,151.4) (88, 177.9) (89, 178.8) (90, 180.5)
		};
			
		\legend{ART, PT, VBT}
		\end{axis}
		\end{tikzpicture}
	\end{center}
	\caption{Run time in checking monotonicity}
	\label{fig:execution-time}
       \vspace{-0.1cm}
\end{figure}

Second,  for RQ2 we have determined the number of test cases generated during testing. 
All three approaches stop once they have detected the first counter example to monotonicity. 
 Figure~\ref{fig:failAtt} shows the number of failed attempts (i.e., generated test cases before finding the first counter example) for each of the classifiers averaging over all the datasets. Our experimental results suggest that VBT always needs the least amount of test cases to testify non-monotonicity. 

Note that our approach has two possible execution instances when not finding counter examples: 
(1) the SMT solver might continously generate counter example candidates which all fail to be real counter examples in the model and 
thus VBT successively retrains the tree until \texttt{MAX\_SAMPLES} candidates have been generated, or 
(2) it immediately stops because the first decision tree is already monotone and the SMT solver finds no counter example at all. 
In the latter case, the number of failed attempts is 0 (which is favourable for a low number of attempts). 
 However, this only occurred in 7 out of the 90 models. 

\pgfplotsset{
	compat=1.5,
	width=\columnwidth,
	height=6cm
}

\begin{figure}[t]
	\begin{center}
		\begin{tikzpicture}
		\begin{axis}[bar width=4,
		ybar,ylabel=no. of failed attempts,
		legend style={at={(0.5,-0.2)},
			anchor=north,legend columns=-1},
		xtick = {1,2,3,4,5,6,7,8,9}, 
		xmin=0, xmax=10,
		ytick = {0,250,500,750,1000}, 
		xticklabels = {kNN, NN, RF, SVM, NB, AB, GB, LR, LBM}]
		\addplot
		[draw=black,pattern=horizontal lines light gray] 
		coordinates
		{(1,42) (2,48) (3,91) (4,43) (5,129) (6,37) (7,115) (8,128) (9, 258)};
		
		\addplot
		[draw=black,pattern=horizontal lines dark gray] 
		coordinates 
		{(1,282) (2,311) (3,382) (4,477) (5,773) (6,480) (7,844) (8,650) (9, 1000)};
		
		\addplot
		[draw=green!40!black, pattern color=gray, pattern=north east lines] 
		coordinates 
		{(1,457) (2,710) (3,880) (4,611) (5,416) (6,683) (7,843) (8,650) (9, 1000)};
		
		\legend{VBT \qquad, ART , PT}
		\end{axis}
		\end{tikzpicture}
		
	\end{center}
	\caption{Number of failed attempts}
	\label{fig:failAtt}
	\vspace*{-.3cm}
\end{figure}

\smallskip
\noindent 
Summarizing the findings of RQ2 in our experiments, we get 

\begin{center}
\fcolorbox{blue}{blue!10}{\parbox{0.85\columnwidth}{On average, VBT is more efficient  
than ART and PT in detecting non-monotonicity of black-box ML models.}}
\end{center}

\bigskip

\noindent For the next two research questions we look at verification-based testing only.

\smallskip 
\textbf{RQ3 - Approximation quality.} Table~\ref{tab:retraining} shows the mean number of re-trainings of the decision tree per 
classifier and data set (mean over 10 runs). For those cases where VBT could not find any non-monotonicity, we have written '-'. 
As the results show, the number of re-trainings is high for the monotonicity aware algorithm LightGBM (LBM), which is consistent with the results shown in Figure~\ref{fig:failAtt}. 
Apart from the classifier, the data set seems to influence the number of retrainings (e.g., Adult needs a large number of re-trainings). 
In general, the numbers are -- however -- relatively low ($\leq$ 10). 
Note that there are also 27 models for which {\em no} retraining at all is needed. 

\smallskip
\noindent 
Hence, we conclude the following. 

\begin{center}
\fcolorbox{blue}{blue!10}{\parbox{0.85\columnwidth}{On average, the approximation quality of decision trees in VBT is good enough 
to only require a small number of retrainings for non-monotonicity detection. }}
\end{center}

\bigskip 
\textbf{RQ4 - Strategy Selection.} For RQ4, we modified VBT as to not stop upon the first valid counter example. 
Figures~\ref{fig:pruneResultsStr} and \ref{fig:pruneResultsWeak} show the {\em detection rates} (number of detected counter examples 
divided by number of test cases) of the two pruning strategies alone in computing strong (Fig.~\ref{fig:pruneResultsStr}) and weak monotonicity (Fig.~\ref{fig:pruneResultsWeak}), respectively. 
Note the difference in the maximal detection rate which is 0.5 for strong and only 0.25 for weak monotonicity. 
On a large number of classifiers branch pruning is better or equal to feature pruning for strong monotonicity 
whereas for weak monotonicity it is the other way round. This can partly be explained on the decision tree itself: since 
weak monotonicity requires all but the values of monotone features to be the same in a pair, branch pruning cannot 
exhibit its full power.

\smallskip
\noindent 
The results suggest the following.  

\begin{center}
\fcolorbox{blue}{blue!10}{\parbox{0.85\columnwidth}{On average, branch pruning achieves a higher detection rate than feature pruning for strong monotonicity and vice versa for weak monotonicity. }}
\end{center}

\subsection{Limitations and Threats to Validity}

Since we employ an SMT solver for monotonicity computation, 
verification-based testing is restricted to feature values allowed by the solver. 
Currently, we have data sets with integer and real values. 
For other domains of feature values, an encoding would be necessary. This is however often done by 
ML algorithms within preprocessing steps anyway, so we could easily make use of existing techniques there.  

Threats to the validity of results are the choice of data sets and ML algorithms and the choice of feature groups for monotonicity checking.
For the ML algorithms, we are confident that we have covered all sorts of basic classifiers in usage today (of course, 
there are in addition numerous specialised classifiers which however often make use of the base techniques). 
As we have taken a number of different, publicly available data sets for machine learning, we are furthermore confident that our 
data sets are diverse enough to exhibit different properties of the approaches and reflect real data sets.  
In particular, the ERA data set exposes interesting properties of verification-based testing. 

A  threat to the internal validity is the high degree of randomness involved in the techniques. First, a number of classifiers use 
randomized algorithms for generating models. 
Thus, in principle we might get one monotone and one non-monotone model when training {\em with the same classifier on exactly the 
same data set}.  
To ensure fairness during comparison, all three approaches were always started with the same model as input (training of MUTs is 
external to testing). 
Second, all three approaches themselves randomly generate (at least some) data instances (ART and PT as potential test cases, 
VBT for oracle data). VBT in addition uses a decision tree training algorithm which itself involves randomness. 
Hence, our decision trees can vary from one run to the next, and this stays so even if we would fix the oracle data. 
To mitigate these threats, all experiments were performed 10 times and the results give the mean over these 10 runs. 

\begin{table}[t]\centering
	
	\small
	\caption{Mean number of re-trainings}
	\label{tab:retraining}
	\begin{tabular}{r r r r r r r r r r}
		\toprule
		\diagbox[width=\dimexpr \textwidth/18+4\tabcolsep\relax, height=1.1cm]{ \textit{Data} }{\textit{Classifier}} & \textbf{kNN} & \textbf{NN} & \textbf{RF} & \textbf{SVM} & \textbf{NB} & \textbf{AB} & \textbf{GB} & \textbf{LR} & \textbf{LBM} \\
		\midrule
		\textbf{Adult} & 4 & 6 & 19 & 7 & 26 & 1 & 25 & 37 & -\\
		\rowcol \textbf{Auto} & 0 & 1 & 3 & 0 & - & 3 & 2 & 0 & 35\\
		\textbf{Car} & 0 & 0 & 0 & 0 & 0 & - & 1 & 5 & -\\
		\rowcol \textbf{CPU} & 0 & 0 & 1 & 0 & 26 & 0 & 9 & 1 & -\\
		\textbf{Diabetes} & 3 & 0 & 1 & 0 & 4 & 8 & 2 & 7 & -\\
		\rowcol \textbf{ERA} & 1 & - & - & - & - & 0 & - & - & -\\
		\textbf{ESL} & - & - & 1 & 0 & - & - & - & - & -\\
		\rowcol \textbf{Housing} & 0 & 0 & 1 & 0 & 1 & 0 & 1 & 0 & -\\
		\textbf{Mammo} & 1 & 0 & 0 & 1 & 1 & 0 & 1 & 5 & -\\
		\rowcol \textbf{Mpg} & 0 & 10 & 0 & 1 & 0 & 0 & 0 & 5 & 18\\
		\bottomrule
	\end{tabular}
	\vspace{0.1cm}
\end{table}

\section{Related work}\label{sec:related}

We divide our discussion of related works in three parts. First, we discuss some works   incorporating monotonicity in the predictive model, then mention some recent techniques for machine learning testing 
and third discuss approaches using model inference in testing.

\paragraph{Generating monotone models} The existing works in monotonicity focus on specific ML algorithms. In~\cite{archer1993application}, Archer et al.~first propose building a monotone neural network model by adjusting the contribution of training samples in the training process. 
There exists some follow up works which constrain the parameters of the neural network algorithm to enforce monotonicity~\cite{DugasBBNG09, Sill97}. In a more recent work You et al.~\cite{YouDCPG17} propose an approach to generate a guaranteedly monotone deep lattice network with respect to a given set of features. 
Lauer et al.~\cite{LauerB08} enforce monotonicity in support vector machines (with linear kernels) by constraining the derivative to be positive within a specified range. Riihimäki et al.~\cite{RiihimakiV10}
build a Gaussian monotone model by using virtual derivative observations. 
In a follow up work, Siivola et al.~\cite{siivola2016automatic} give an approach based on the same idea to detect monotonicity only for Gaussian distributions. 
Although using derivatives of the function can work for some ML algorithms, it cannot be generalized and is not possible to use in algorithms where the learned functions (i.e., models) are non-linear.



\pgfplotsset{
	compat=1.5,
	width=\columnwidth,
	height=5cm
}

\begin{figure}[t]
	\begin{center}
		\begin{tikzpicture}
		\begin{axis}[bar width=5,
		ybar,ylabel=detection rate,
		legend style={at={(0.5,-0.2)},
			anchor=north,legend columns=-1},
		xtick = {1,2,3,4,5,6,7,8}, 
		xmin=0, xmax=9,
		ytick = {0.0,0.25,0.50,0.75,1.0}, 
             yticklabels = {0.0,0.25,0.50,0.75,1.0}, 
		xticklabels = {kNN, NN, RF, SVM, NB, AB, GB, LR}]
		\addplot
		[draw=gray,pattern=horizontal lines light gray] 
		coordinates
		{(1,0.35) (2,0.3) (3,0.31) (4,0.3) (5,0.34) (6,0.5) (7,0.34) (8,0.30)};
		
		\addplot
		[draw=gray,pattern=horizontal lines dark gray] 
		coordinates 
		{(1,0.37) (2,0.33) (3,0.32) (4,0.3) (5,0.33) (6,0.5) (7,0.4) (8,0.3)};

		\legend{Feature-prune \qquad, Branch-prune}
		\end{axis}
		\end{tikzpicture}
		
	\end{center}
	\caption{Performance of branch and feature pruning in computing strong monotonicity}
	\label{fig:pruneResultsStr}
	\vspace*{-.3cm}
\end{figure}

\paragraph{Validating models} There are a number of recent works which aim at validating properties of predictive models, none of which however have looked at monotonicity. 
In~\cite{HuangKWW17}, Huang et al.~propose {\em robustness} as a safety property and give a verification technique showing that a Deep Neural Network (DNN) guarantees postconditions to hold on its outputs when the inputs satisfy a given precondition. 
Gehr et al.~\cite{GehrMDTCV18} use abstract interpretation to verify robustness of DNN models. Pei et al.~\cite{PeiCYJ17} propose the first white-box testing technique to test DNNs. They use neuron coverage as a criterion to generate the test cases for the predictive model. 
Sun et al.~\cite{SunWRHKK18} propose concolic testing to test the robustness property of DNNs. The authors use a set of coverage requirements (such as neuron, MC/DC and neuron boundary coverage, Lipschitz continuity) to generate test inputs. 

Recently, Sharma et al.~\cite{SharmaW19} have proposed a property called {\em balancedness} on the learning algorithm. They perform specific transformations on the training data and check whether the learning algorithm generates a different predictive model after applying such transformations. This work thus focusses on testing the ML algorithm itself, not the model. 

In~\cite{galhotra2017fairness}, Galhotra et al.~perform black-box testing to check {\em fairness} of the predictive model. Basically, they use random testing with confidence driven sampling.
 It has the drawback of generating completely random sets of test data without considering the structure of the model. 

The work closest to us is that of Agarwal et al.~\cite{AggarwalLNDS19}. They also study fairness testing, but compared to Galhotra et al.~\cite{galhotra2017fairness} they aim at a more systematic generation 
of test inputs. To this end, they employ LIME~\cite{Ribeiro0G16} (a tool for generating {\em local} explanations for predictions) to generate a {\em partial} decision tree (often just a path in the tree) from the black-box model. On this path, they use dynamic symbolic execution to generate multiple test cases, much alike we do. The difference to our work is that we generate a 
decision tree approximating the {\em entire} black-box model under test, and -- more importantly -- 
we use the generated white-box model for the {\em computation} of test inputs (potential 
counter examples to monotonicity). We thus achieve a targeted test case generation.
 
%


\begin{figure}[t]
	\begin{center}
		\begin{tikzpicture}
		\begin{axis}[bar width=5,
		ybar,ylabel=detection rate,
		legend style={at={(0.5,-0.2)},
			anchor=north,legend columns=-1},
		xtick = {1,2,3,4,5,6,7,8}, 
		xmin=0, xmax=9,
		ytick = {0.0,0.25,0.50}, 
		xticklabels = {kNN, NN, RF, SVM, NB, AB, GB, LR}]
		
		\addplot
		[draw=gray,pattern=horizontal lines light gray] 
		coordinates 
		{(1,0.25) (2,0.21) (3,0.17) (4,0.15) (5,0.09) (6,0.23) (7,0.2) (8,0.07)};
		
		\addplot
		[draw=gray,pattern=horizontal lines dark gray] 
		coordinates
		{(1,0.2) (2,0.16) (3,0.17) (4,0.15) (5,0.1) (6,0.25) (7,0.18) (8,0.05)};
		
		\legend{Feature-prune \qquad, Branch-prune}
		\end{axis}
		\end{tikzpicture}
		
	\end{center}
	\caption{Performance of branch and feature pruning in computing weak monotonicity}
	\label{fig:pruneResultsWeak}
	\vspace*{-.3cm}
\end{figure}

\paragraph{Testing via model inference}

The use of learning in testing has long been considered in the field of model-based testing. 
Therein, learning is used to extract a model of the system under test. 
Such models most often are some sort of automaton (finite state machine) and learning is based on Angluin's L$^*$ algorithm~\cite{DBLP:journals/iandc/Angluin87}. 
For a survey of techniques see~\cite{DBLP:conf/dagstuhl/AichernigMMTT16,DBLP:conf/dagstuhl/Meinke16}. 

In contrast to this, we employ machine learning techniques to infer a model. 
The inference of a decision tree describing the behaviour of software has already been pursued 
by Papadopoulus and Walkinshaw~\cite{DBLP:conf/icse/PapadopoulosW15} as well as Briand et al.~\cite{DBLP:journals/infsof/BriandLBS09}. 
The former -- similar to us -- translate the decision tree to logic in order to have Z3 generate 
test inputs covering different branches of the tree. However, they do not employ the tree to 
generate counter examples to the property to be tested. 
Thus, the advantage of having a verifiable white-box model for targeted test input generation 
is not utilized. Briand et al.~on the other hand use the decision tree in a semi-automated 
approach to the re-engineering of test suites. This approach requires the manual inspection of 
the decision tree by testers. 
A survey on inference-driven techniques is given in~\cite{DBLP:conf/dagstuhl/Walkinshaw16}. 

%
\section{Conclusion}\label{sec:conclusion}

In this work, we have defined the property of monotonicity of ML models and have proposed a novel approach to testing monotonicity. Our technique approximates the black-box model by a white-box model  and applies SMT solving techniques to compute monotonicity on the white-box model.
We have evaluated the effectiveness and efficiency of our approach by applying it to several ML models and found our approach to outperform 
both adaptive random and property-based testing. 

As future work, we plan to apply this scheme to validate other important properties of ML models. 
Our white-box model easily allows for  checking other properties, like for instance fairness, just by
 applying a different check on the generated SMT code. 
Also, we would like to improve our framework by using incremental learning to avoid re-training 
of the entire decision tree.

\bibliographystyle{ACM-Reference-Format}
\bibliography{ref}


\begin{thebibliography}{36}


\ifx \showCODEN    \undefined \def \showCODEN     #1{\unskip}     \fi
\ifx \showDOI      \undefined \def \showDOI       #1{#1}\fi
\ifx \showISBNx    \undefined \def \showISBNx     #1{\unskip}     \fi
\ifx \showISBNxiii \undefined \def \showISBNxiii  #1{\unskip}     \fi
\ifx \showISSN     \undefined \def \showISSN      #1{\unskip}     \fi
\ifx \showLCCN     \undefined \def \showLCCN      #1{\unskip}     \fi
\ifx \shownote     \undefined \def \shownote      #1{#1}          \fi
\ifx \showarticletitle \undefined \def \showarticletitle #1{#1}   \fi
\ifx \showURL      \undefined \def \showURL       {\relax}        \fi
\providecommand\bibfield[2]{#2}
\providecommand\bibinfo[2]{#2}
\providecommand\natexlab[1]{#1}
\providecommand\showeprint[2][]{arXiv:#2}

\bibitem[\protect\citeauthoryear{??}{hyp}{2019}]%
        {hypothesis}
 \bibinfo{year}{2019}\natexlab{}.
\newblock \bibinfo{title}{Hypothesis}.
\newblock
  \bibinfo{howpublished}{\url{https://github.com/HypothesisWorks/hypothesis}}.
\newblock


\bibitem[\protect\citeauthoryear{??}{lig}{2019}]%
        {light}
 \bibinfo{year}{2019}\natexlab{}.
\newblock \bibinfo{title}{LightGBM}.
\newblock \bibinfo{howpublished}{\url{https://github.com/Microsoft/LightGBM}}.
\newblock


\bibitem[\protect\citeauthoryear{Aggarwal, Lohia, Nagar, Dey, and
  Saha}{Aggarwal et~al\mbox{.}}{2019}]%
        {AggarwalLNDS19}
\bibfield{author}{\bibinfo{person}{Aniya Aggarwal}, \bibinfo{person}{Pranay
  Lohia}, \bibinfo{person}{Seema Nagar}, \bibinfo{person}{Kuntal Dey}, {and}
  \bibinfo{person}{Diptikalyan Saha}.} \bibinfo{year}{2019}\natexlab{}.
\newblock \showarticletitle{Black box fairness testing of machine learning
  models}. In \bibinfo{booktitle}{\emph{Proceedings of the {ACM} Joint Meeting
  on European Software Engineering Conference and Symposium on the Foundations
  of Software Engineering, {ESEC/SIGSOFT} {FSE}}}. \bibinfo{pages}{625--635}.
\newblock
\urldef\tempurl%
\url{https://doi.org/10.1145/3338906.3338937}
\showDOI{\tempurl}


\bibitem[\protect\citeauthoryear{Aichernig, Mostowski, Mousavi, Tappler, and
  Taromirad}{Aichernig et~al\mbox{.}}{2018}]%
        {DBLP:conf/dagstuhl/AichernigMMTT16}
\bibfield{author}{\bibinfo{person}{Bernhard~K. Aichernig},
  \bibinfo{person}{Wojciech Mostowski}, \bibinfo{person}{Mohammad~Reza
  Mousavi}, \bibinfo{person}{Martin Tappler}, {and} \bibinfo{person}{Masoumeh
  Taromirad}.} \bibinfo{year}{2018}\natexlab{}.
\newblock \showarticletitle{Model Learning and Model-Based Testing}, See
  \citeN{DBLP:conf/dagstuhl/}, \bibinfo{pages}{74--100}.
\newblock
\showISBNx{978-3-319-96561-1}
\urldef\tempurl%
\url{https://doi.org/10.1007/978-3-319-96562-8\_3}
\showDOI{\tempurl}


\bibitem[\protect\citeauthoryear{Angluin}{Angluin}{1987}]%
        {DBLP:journals/iandc/Angluin87}
\bibfield{author}{\bibinfo{person}{Dana Angluin}.}
  \bibinfo{year}{1987}\natexlab{}.
\newblock \showarticletitle{Learning Regular Sets from Queries and
  Counterexamples}.
\newblock \bibinfo{journal}{\emph{Inf. Comput.}} \bibinfo{volume}{75},
  \bibinfo{number}{2} (\bibinfo{year}{1987}), \bibinfo{pages}{87--106}.
\newblock
\urldef\tempurl%
\url{https://doi.org/10.1016/0890-5401(87)90052-6}
\showDOI{\tempurl}


\bibitem[\protect\citeauthoryear{Archer and Wang}{Archer and Wang}{1993}]%
        {archer1993application}
\bibfield{author}{\bibinfo{person}{Norman~P Archer} {and}
  \bibinfo{person}{Shouhong Wang}.} \bibinfo{year}{1993}\natexlab{}.
\newblock \showarticletitle{Application of the back propagation neural network
  algorithm with monotonicity constraints for two-group classification
  problems}.
\newblock \bibinfo{journal}{\emph{Decision Sciences}} \bibinfo{volume}{24},
  \bibinfo{number}{1} (\bibinfo{year}{1993}), \bibinfo{pages}{60--75}.
\newblock


\bibitem[\protect\citeauthoryear{Bennaceur, H{\"{a}}hnle, and Meinke}{Bennaceur
  et~al\mbox{.}}{2018}]%
        {DBLP:conf/dagstuhl/}
\bibfield{editor}{\bibinfo{person}{Amel Bennaceur}, \bibinfo{person}{Reiner
  H{\"{a}}hnle}, {and} \bibinfo{person}{Karl Meinke}} (Eds.).
  \bibinfo{year}{2018}\natexlab{}.
\newblock \bibinfo{booktitle}{\emph{Machine Learning for Dynamic Software
  Analysis}}. \bibinfo{series}{Lecture Notes in Computer Science},
  Vol.~\bibinfo{volume}{11026}. \bibinfo{publisher}{Springer}.
\newblock
\showISBNx{978-3-319-96561-1}
\urldef\tempurl%
\url{https://doi.org/10.1007/978-3-319-96562-8}
\showDOI{\tempurl}


\bibitem[\protect\citeauthoryear{Briand, Labiche, Bawar, and Spido}{Briand
  et~al\mbox{.}}{2009}]%
        {DBLP:journals/infsof/BriandLBS09}
\bibfield{author}{\bibinfo{person}{Lionel~C. Briand}, \bibinfo{person}{Yvan
  Labiche}, \bibinfo{person}{Zaheer Bawar}, {and} \bibinfo{person}{Nadia~Traldi
  Spido}.} \bibinfo{year}{2009}\natexlab{}.
\newblock \showarticletitle{Using machine learning to refine Category-Partition
  test specifications and test suites}.
\newblock \bibinfo{journal}{\emph{Information {\&} Software Technology}}
  \bibinfo{volume}{51}, \bibinfo{number}{11} (\bibinfo{year}{2009}),
  \bibinfo{pages}{1551--1564}.
\newblock
\urldef\tempurl%
\url{https://doi.org/10.1016/j.infsof.2009.06.006}
\showDOI{\tempurl}


\bibitem[\protect\citeauthoryear{Carlini and Wagner}{Carlini and
  Wagner}{2017}]%
        {Carlini017}
\bibfield{author}{\bibinfo{person}{Nicholas Carlini} {and}
  \bibinfo{person}{David~A. Wagner}.} \bibinfo{year}{2017}\natexlab{}.
\newblock \showarticletitle{Towards Evaluating the Robustness of Neural
  Networks}. In \bibinfo{booktitle}{\emph{2017 {IEEE} Symposium on Security and
  Privacy, {SP}}}. \bibinfo{pages}{39--57}.
\newblock
\urldef\tempurl%
\url{https://doi.org/10.1109/SP.2017.49}
\showDOI{\tempurl}


\bibitem[\protect\citeauthoryear{Chen, Leung, and Mak}{Chen
  et~al\mbox{.}}{2004}]%
        {DBLP:conf/asian/ChenLM04}
\bibfield{author}{\bibinfo{person}{Tsong~Yueh Chen}, \bibinfo{person}{Hing
  Leung}, {and} \bibinfo{person}{I.~K. Mak}.} \bibinfo{year}{2004}\natexlab{}.
\newblock \showarticletitle{Adaptive Random Testing}. In
  \bibinfo{booktitle}{\emph{{ASIAN}}} \emph{(\bibinfo{series}{Lecture Notes in
  Computer Science})}, \bibfield{editor}{\bibinfo{person}{Michael~J. Maher}}
  (Ed.), Vol.~\bibinfo{volume}{3321}. \bibinfo{publisher}{Springer},
  \bibinfo{pages}{320--329}.
\newblock
\showISBNx{3-540-24087-X}
\urldef\tempurl%
\url{https://doi.org/10.1007/978-3-540-30502-6\_23}
\showDOI{\tempurl}


\bibitem[\protect\citeauthoryear{Claessen and Hughes}{Claessen and
  Hughes}{2000}]%
        {DBLP:conf/icfp/ClaessenH00}
\bibfield{author}{\bibinfo{person}{Koen Claessen} {and} \bibinfo{person}{John
  Hughes}.} \bibinfo{year}{2000}\natexlab{}.
\newblock \showarticletitle{QuickCheck: a lightweight tool for random testing
  of Haskell programs}. In \bibinfo{booktitle}{\emph{{(ICFP} '00)}},
  \bibfield{editor}{\bibinfo{person}{Martin Odersky} {and}
  \bibinfo{person}{Philip Wadler}} (Eds.). \bibinfo{publisher}{{ACM}},
  \bibinfo{pages}{268--279}.
\newblock
\showISBNx{1-58113-202-6}
\urldef\tempurl%
\url{https://doi.org/10.1145/351240.351266}
\showDOI{\tempurl}


\bibitem[\protect\citeauthoryear{de~Moura and Bj{\o}rner}{de~Moura and
  Bj{\o}rner}{2008}]%
        {MouraB08}
\bibfield{author}{\bibinfo{person}{Leonardo~Mendon{\c{c}}a de Moura} {and}
  \bibinfo{person}{Nikolaj Bj{\o}rner}.} \bibinfo{year}{2008}\natexlab{}.
\newblock \showarticletitle{{Z3:} An Efficient {SMT} Solver}. In
  \bibinfo{booktitle}{\emph{Tools and Algorithms for the Construction and
  Analysis of Systems, 14th International Conference, {TACAS} 2008}}.
  \bibinfo{pages}{337--340}.
\newblock
\urldef\tempurl%
\url{https://doi.org/10.1007/978-3-540-78800-3\_24}
\showDOI{\tempurl}


\bibitem[\protect\citeauthoryear{Dugas, Bengio, B{\'{e}}lisle, Nadeau, and
  Garcia}{Dugas et~al\mbox{.}}{2009}]%
        {DugasBBNG09}
\bibfield{author}{\bibinfo{person}{Charles Dugas}, \bibinfo{person}{Yoshua
  Bengio}, \bibinfo{person}{Fran{\c{c}}ois B{\'{e}}lisle},
  \bibinfo{person}{Claude Nadeau}, {and} \bibinfo{person}{Ren{\'{e}} Garcia}.}
  \bibinfo{year}{2009}\natexlab{}.
\newblock \showarticletitle{Incorporating Functional Knowledge in Neural
  Networks}.
\newblock \bibinfo{journal}{\emph{J. Mach. Learn. Res.}}  \bibinfo{volume}{10}
  (\bibinfo{year}{2009}), \bibinfo{pages}{1239--1262}.
\newblock
\urldef\tempurl%
\url{https://dl.acm.org/citation.cfm?id=1577111}
\showURL{%
\tempurl}


\bibitem[\protect\citeauthoryear{Galhotra, Brun, and Meliou}{Galhotra
  et~al\mbox{.}}{2017}]%
        {galhotra2017fairness}
\bibfield{author}{\bibinfo{person}{Sainyam Galhotra}, \bibinfo{person}{Yuriy
  Brun}, {and} \bibinfo{person}{Alexandra Meliou}.}
  \bibinfo{year}{2017}\natexlab{}.
\newblock \showarticletitle{Fairness testing: testing software for
  discrimination}. In \bibinfo{booktitle}{\emph{Proceedings of the 2017 11th
  Joint Meeting on Foundations of Software Engineering}}. ACM,
  \bibinfo{pages}{498--510}.
\newblock


\bibitem[\protect\citeauthoryear{Gehr, Mirman, Drachsler{-}Cohen, Tsankov,
  Chaudhuri, and Vechev}{Gehr et~al\mbox{.}}{2018}]%
        {GehrMDTCV18}
\bibfield{author}{\bibinfo{person}{Timon Gehr}, \bibinfo{person}{Matthew
  Mirman}, \bibinfo{person}{Dana Drachsler{-}Cohen}, \bibinfo{person}{Petar
  Tsankov}, \bibinfo{person}{Swarat Chaudhuri}, {and}
  \bibinfo{person}{Martin~T. Vechev}.} \bibinfo{year}{2018}\natexlab{}.
\newblock \showarticletitle{{AI2:} Safety and Robustness Certification of
  Neural Networks with Abstract Interpretation}. In
  \bibinfo{booktitle}{\emph{{IEEE} Symposium on Security and Privacy, {SP}}}.
  \bibinfo{pages}{3--18}.
\newblock
\urldef\tempurl%
\url{https://doi.org/10.1109/SP.2018.00058}
\showDOI{\tempurl}


\bibitem[\protect\citeauthoryear{Godefroid, Klarlund, and Sen}{Godefroid
  et~al\mbox{.}}{2005}]%
        {GodefroidKS05}
\bibfield{author}{\bibinfo{person}{Patrice Godefroid}, \bibinfo{person}{Nils
  Klarlund}, {and} \bibinfo{person}{Koushik Sen}.}
  \bibinfo{year}{2005}\natexlab{}.
\newblock \showarticletitle{{DART:} directed automated random testing}. In
  \bibinfo{booktitle}{\emph{Proceedings of the {ACM} {SIGPLAN} 2005 Conference
  on Programming Language Design and Implementation}}.
  \bibinfo{pages}{213--223}.
\newblock
\urldef\tempurl%
\url{https://doi.org/10.1145/1065010.1065036}
\showDOI{\tempurl}


\bibitem[\protect\citeauthoryear{Guidotti, Monreale, Ruggieri, Turini,
  Giannotti, and Pedreschi}{Guidotti et~al\mbox{.}}{2019}]%
        {GuidottiMRTGP19}
\bibfield{author}{\bibinfo{person}{Riccardo Guidotti}, \bibinfo{person}{Anna
  Monreale}, \bibinfo{person}{Salvatore Ruggieri}, \bibinfo{person}{Franco
  Turini}, \bibinfo{person}{Fosca Giannotti}, {and} \bibinfo{person}{Dino
  Pedreschi}.} \bibinfo{year}{2019}\natexlab{}.
\newblock \showarticletitle{A Survey of Methods for Explaining Black Box
  Models}.
\newblock \bibinfo{journal}{\emph{{ACM} Comput. Surv.}} \bibinfo{volume}{51},
  \bibinfo{number}{5} (\bibinfo{year}{2019}), \bibinfo{pages}{93:1--93:42}.
\newblock
\urldef\tempurl%
\url{https://doi.org/10.1145/3236009}
\showDOI{\tempurl}


\bibitem[\protect\citeauthoryear{Huang, Kwiatkowska, Wang, and Wu}{Huang
  et~al\mbox{.}}{2017}]%
        {HuangKWW17}
\bibfield{author}{\bibinfo{person}{Xiaowei Huang}, \bibinfo{person}{Marta
  Kwiatkowska}, \bibinfo{person}{Sen Wang}, {and} \bibinfo{person}{Min Wu}.}
  \bibinfo{year}{2017}\natexlab{}.
\newblock \showarticletitle{Safety Verification of Deep Neural Networks}. In
  \bibinfo{booktitle}{\emph{Computer Aided Verification - 29th International
  Conference, {CAV}}}. \bibinfo{pages}{3--29}.
\newblock
\urldef\tempurl%
\url{https://doi.org/10.1007/978-3-319-63387-9\_1}
\showDOI{\tempurl}


\bibitem[\protect\citeauthoryear{Johansson and Niklasson}{Johansson and
  Niklasson}{[n.d.]}]%
        {JohanssonN09}
\bibfield{author}{\bibinfo{person}{Ulf Johansson} {and} \bibinfo{person}{Lars
  Niklasson}.} \bibinfo{year}{[n.d.]}\natexlab{}.
\newblock \showarticletitle{Evolving decision trees using oracle guides}. In
  \bibinfo{booktitle}{\emph{Proceedings of the {IEEE} Symposium on
  Computational Intelligence and Data Mining, {CIDM} 2009}}.
  \bibinfo{pages}{238--244}.
\newblock
\urldef\tempurl%
\url{https://doi.org/10.1109/CIDM.2009.4938655}
\showDOI{\tempurl}


\bibitem[\protect\citeauthoryear{King}{King}{1976}]%
        {King76}
\bibfield{author}{\bibinfo{person}{James~C. King}.}
  \bibinfo{year}{1976}\natexlab{}.
\newblock \showarticletitle{Symbolic Execution and Program Testing}.
\newblock \bibinfo{journal}{\emph{Commun. {ACM}}} \bibinfo{volume}{19},
  \bibinfo{number}{7} (\bibinfo{year}{1976}), \bibinfo{pages}{385--394}.
\newblock
\urldef\tempurl%
\url{https://doi.org/10.1145/360248.360252}
\showDOI{\tempurl}


\bibitem[\protect\citeauthoryear{Kotlowski and Slowinski}{Kotlowski and
  Slowinski}{2009}]%
        {KotlowskiS09}
\bibfield{author}{\bibinfo{person}{Wojciech Kotlowski} {and}
  \bibinfo{person}{Roman Slowinski}.} \bibinfo{year}{2009}\natexlab{}.
\newblock \showarticletitle{Rule learning with monotonicity constraints}. In
  \bibinfo{booktitle}{\emph{Proceedings of the 26th Annual International
  Conference on Machine Learning, {ICML} 2009}}. \bibinfo{pages}{537--544}.
\newblock
\urldef\tempurl%
\url{https://doi.org/10.1145/1553374.1553444}
\showDOI{\tempurl}


\bibitem[\protect\citeauthoryear{Lauer and Bloch}{Lauer and Bloch}{2008}]%
        {LauerB08}
\bibfield{author}{\bibinfo{person}{Fabien Lauer} {and}
  \bibinfo{person}{G{\'{e}}rard Bloch}.} \bibinfo{year}{2008}\natexlab{}.
\newblock \showarticletitle{Incorporating prior knowledge in support vector
  regression}.
\newblock \bibinfo{journal}{\emph{Machine Learning}} \bibinfo{volume}{70},
  \bibinfo{number}{1} (\bibinfo{year}{2008}), \bibinfo{pages}{89--118}.
\newblock
\urldef\tempurl%
\url{https://doi.org/10.1007/s10994-007-5035-5}
\showDOI{\tempurl}


\bibitem[\protect\citeauthoryear{Meinke}{Meinke}{2018}]%
        {DBLP:conf/dagstuhl/Meinke16}
\bibfield{author}{\bibinfo{person}{Karl Meinke}.}
  \bibinfo{year}{2018}\natexlab{}.
\newblock \showarticletitle{Learning-Based Testing: Recent Progress and Future
  Prospects}, See \citeN{DBLP:conf/dagstuhl/}, \bibinfo{pages}{53--73}.
\newblock
\showISBNx{978-3-319-96561-1}
\urldef\tempurl%
\url{https://doi.org/10.1007/978-3-319-96562-8\_2}
\showDOI{\tempurl}


\bibitem[\protect\citeauthoryear{Papadopoulos and Walkinshaw}{Papadopoulos and
  Walkinshaw}{2015}]%
        {DBLP:conf/icse/PapadopoulosW15}
\bibfield{author}{\bibinfo{person}{Petros Papadopoulos} {and}
  \bibinfo{person}{Neil Walkinshaw}.} \bibinfo{year}{2015}\natexlab{}.
\newblock \showarticletitle{Black-Box Test Generation from Inferred Models}. In
  \bibinfo{booktitle}{\emph{{RAISE}}},
  \bibfield{editor}{\bibinfo{person}{Rachel Harrison},
  \bibinfo{person}{Ayse~Basar Bener}, {and} \bibinfo{person}{Burak Turhan}}
  (Eds.). \bibinfo{publisher}{{IEEE} Computer Society},
  \bibinfo{pages}{19--24}.
\newblock
\showISBNx{978-1-4673-7064-6}
\urldef\tempurl%
\url{https://doi.org/10.1109/RAISE.2015.11}
\showDOI{\tempurl}


\bibitem[\protect\citeauthoryear{Pei, Cao, Yang, and Jana}{Pei
  et~al\mbox{.}}{2017}]%
        {PeiCYJ17}
\bibfield{author}{\bibinfo{person}{Kexin Pei}, \bibinfo{person}{Yinzhi Cao},
  \bibinfo{person}{Junfeng Yang}, {and} \bibinfo{person}{Suman Jana}.}
  \bibinfo{year}{2017}\natexlab{}.
\newblock \showarticletitle{DeepXplore: Automated Whitebox Testing of Deep
  Learning Systems}. In \bibinfo{booktitle}{\emph{Proceedings of the 26th
  Symposium on Operating Systems Principles,}}. \bibinfo{pages}{1--18}.
\newblock
\urldef\tempurl%
\url{https://doi.org/10.1145/3132747.3132785}
\showDOI{\tempurl}


\bibitem[\protect\citeauthoryear{Potharst and Feelders}{Potharst and
  Feelders}{2002}]%
        {Potharst:2002:CTP:568574.568577}
\bibfield{author}{\bibinfo{person}{R. Potharst} {and} \bibinfo{person}{A.~J.
  Feelders}.} \bibinfo{year}{2002}\natexlab{}.
\newblock \showarticletitle{Classification Trees for Problems with Monotonicity
  Constraints}.
\newblock \bibinfo{journal}{\emph{SIGKDD Explor. Newsl.}} \bibinfo{volume}{4},
  \bibinfo{number}{1} (\bibinfo{date}{June} \bibinfo{year}{2002}),
  \bibinfo{pages}{1--10}.
\newblock
\showISSN{1931-0145}
\urldef\tempurl%
\url{https://doi.org/10.1145/568574.568577}
\showDOI{\tempurl}


\bibitem[\protect\citeauthoryear{Ribeiro, Singh, and Guestrin}{Ribeiro
  et~al\mbox{.}}{2016}]%
        {Ribeiro0G16}
\bibfield{author}{\bibinfo{person}{Marco~T{\'{u}}lio Ribeiro},
  \bibinfo{person}{Sameer Singh}, {and} \bibinfo{person}{Carlos Guestrin}.}
  \bibinfo{year}{2016}\natexlab{}.
\newblock \showarticletitle{``\text{Why} Should {I} Trust You?": Explaining the
  Predictions of Any Classifier}. In \bibinfo{booktitle}{\emph{Proceedings of
  the 22nd {ACM} {SIGKDD} International Conference on Knowledge Discovery and
  Data Mining}}. \bibinfo{pages}{1135--1144}.
\newblock
\urldef\tempurl%
\url{https://doi.org/10.1145/2939672.2939778}
\showDOI{\tempurl}


\bibitem[\protect\citeauthoryear{Riihim{\"{a}}ki and Vehtari}{Riihim{\"{a}}ki
  and Vehtari}{2010}]%
        {RiihimakiV10}
\bibfield{author}{\bibinfo{person}{Jaakko Riihim{\"{a}}ki} {and}
  \bibinfo{person}{Aki Vehtari}.} \bibinfo{year}{2010}\natexlab{}.
\newblock \showarticletitle{Gaussian processes with monotonicity information}.
  In \bibinfo{booktitle}{\emph{Proceedings of the Thirteenth International
  Conference on Artificial Intelligence and Statistics, {AISTATS}}}.
  \bibinfo{pages}{645--652}.
\newblock
\urldef\tempurl%
\url{http://proceedings.mlr.press/v9/riihimaki10a.html}
\showURL{%
\tempurl}


\bibitem[\protect\citeauthoryear{Sen, Marinov, and Agha}{Sen
  et~al\mbox{.}}{2005}]%
        {SenMA05}
\bibfield{author}{\bibinfo{person}{Koushik Sen}, \bibinfo{person}{Darko
  Marinov}, {and} \bibinfo{person}{Gul Agha}.} \bibinfo{year}{2005}\natexlab{}.
\newblock \showarticletitle{{CUTE:} a concolic unit testing engine for {C}}. In
  \bibinfo{booktitle}{\emph{Proceedings of the 10th European Software
  Engineering Conference held jointly with 13th {ACM} {SIGSOFT} International
  Symposium on Foundations of Software Engineering}}.
  \bibinfo{pages}{263--272}.
\newblock
\urldef\tempurl%
\url{https://doi.org/10.1145/1081706.1081750}
\showDOI{\tempurl}


\bibitem[\protect\citeauthoryear{Sharma and Wehrheim}{Sharma and
  Wehrheim}{2019}]%
        {SharmaW19}
\bibfield{author}{\bibinfo{person}{Arnab Sharma} {and} \bibinfo{person}{Heike
  Wehrheim}.} \bibinfo{year}{2019}\natexlab{}.
\newblock \showarticletitle{Testing Machine Learning Algorithms for Balanced
  Data Usage}. In \bibinfo{booktitle}{\emph{12th {IEEE} Conference on Software
  Testing, Validation and Verification, {ICST}}}. \bibinfo{pages}{125--135}.
\newblock
\urldef\tempurl%
\url{https://doi.org/10.1109/ICST.2019.00022}
\showURL{%
\tempurl}


\bibitem[\protect\citeauthoryear{Siivola, Piironen, and Vehtari}{Siivola
  et~al\mbox{.}}{2016}]%
        {siivola2016automatic}
\bibfield{author}{\bibinfo{person}{Eero Siivola}, \bibinfo{person}{Juho
  Piironen}, {and} \bibinfo{person}{Aki Vehtari}.}
  \bibinfo{year}{2016}\natexlab{}.
\newblock \showarticletitle{Automatic monotonicity detection for Gaussian
  Processes}.
\newblock \bibinfo{journal}{\emph{arXiv preprint arXiv:1610.05440}}
  (\bibinfo{year}{2016}).
\newblock


\bibitem[\protect\citeauthoryear{Sill}{Sill}{1997}]%
        {Sill97}
\bibfield{author}{\bibinfo{person}{Joseph Sill}.}
  \bibinfo{year}{1997}\natexlab{}.
\newblock \showarticletitle{Monotonic Networks}. In
  \bibinfo{booktitle}{\emph{Advances in Neural Information Processing
  Systems}}. \bibinfo{pages}{661--667}.
\newblock
\urldef\tempurl%
\url{http://papers.nips.cc/paper/1358-monotonic-networks}
\showURL{%
\tempurl}


\bibitem[\protect\citeauthoryear{Sun, Wu, Ruan, Huang, Kwiatkowska, and
  Kroening}{Sun et~al\mbox{.}}{2018}]%
        {SunWRHKK18}
\bibfield{author}{\bibinfo{person}{Youcheng Sun}, \bibinfo{person}{Min Wu},
  \bibinfo{person}{Wenjie Ruan}, \bibinfo{person}{Xiaowei Huang},
  \bibinfo{person}{Marta Kwiatkowska}, {and} \bibinfo{person}{Daniel
  Kroening}.} \bibinfo{year}{2018}\natexlab{}.
\newblock \showarticletitle{Concolic testing for deep neural networks}. In
  \bibinfo{booktitle}{\emph{Proceedings of the 33rd {ACM/IEEE} International
  Conference on Automated Software Engineering, {ASE}}}.
  \bibinfo{pages}{109--119}.
\newblock
\urldef\tempurl%
\url{https://doi.org/10.1145/3238147.3238172}
\showDOI{\tempurl}


\bibitem[\protect\citeauthoryear{Tehrani, Cheng, Dembczynski, and
  H{\"{u}}llermeier}{Tehrani et~al\mbox{.}}{2011}]%
        {TehraniCDH11}
\bibfield{author}{\bibinfo{person}{Ali~Fallah Tehrani}, \bibinfo{person}{Weiwei
  Cheng}, \bibinfo{person}{Krzysztof Dembczynski}, {and} \bibinfo{person}{Eyke
  H{\"{u}}llermeier}.} \bibinfo{year}{2011}\natexlab{}.
\newblock \showarticletitle{Learning Monotone Nonlinear Models Using the
  Choquet Integral}. In \bibinfo{booktitle}{\emph{Machine Learning and
  Knowledge Discovery in Databases - European Conference, {ECML}}}.
  \bibinfo{pages}{414--429}.
\newblock
\urldef\tempurl%
\url{https://doi.org/10.1007/978-3-642-23808-6\_27}
\showDOI{\tempurl}


\bibitem[\protect\citeauthoryear{Walkinshaw}{Walkinshaw}{2018}]%
        {DBLP:conf/dagstuhl/Walkinshaw16}
\bibfield{author}{\bibinfo{person}{Neil Walkinshaw}.}
  \bibinfo{year}{2018}\natexlab{}.
\newblock \showarticletitle{Testing Functional Black-Box Programs Without a
  Specification}, See \citeN{DBLP:conf/dagstuhl/}, \bibinfo{pages}{101--120}.
\newblock
\showISBNx{978-3-319-96561-1}
\urldef\tempurl%
\url{https://doi.org/10.1007/978-3-319-96562-8\_4}
\showDOI{\tempurl}


\bibitem[\protect\citeauthoryear{You, Ding, Canini, Pfeifer, and Gupta}{You
  et~al\mbox{.}}{2017}]%
        {YouDCPG17}
\bibfield{author}{\bibinfo{person}{Seungil You}, \bibinfo{person}{David Ding},
  \bibinfo{person}{Kevin~Robert Canini}, \bibinfo{person}{Jan Pfeifer}, {and}
  \bibinfo{person}{Maya~R. Gupta}.} \bibinfo{year}{2017}\natexlab{}.
\newblock \showarticletitle{Deep Lattice Networks and Partial Monotonic
  Functions}. In \bibinfo{booktitle}{\emph{Advances in Neural Information
  Processing Systems 30: Annual Conference on Neural Information Processing
  Systems}}. \bibinfo{pages}{2985--2993}.
\newblock
\urldef\tempurl%
\url{http://papers.nips.cc/paper/6891-deep-lattice-networks-and-partial-monotonic-functions}
\showURL{%
\tempurl}


\end{thebibliography}

\end{document}